%% file: neurips_2026.tex
\documentclass{article}

\PassOptionsToPackage{numbers,compress}{natbib}

\usepackage[main, final]{neurips_2026}

\makeatletter
\renewcommand{\@notice}{}
\makeatother

\usepackage[utf8]{inputenc}
\usepackage[T1]{fontenc}
\usepackage[hypertexnames=false]{hyperref}
\usepackage{url}
\usepackage{subcaption}
\usepackage{booktabs}
\usepackage{amsfonts}
\usepackage{amsmath}
\usepackage{nicefrac}
\usepackage{microtype}
\usepackage{xcolor}
\usepackage{xspace}
\usepackage{graphicx}
\usepackage{multirow}
\usepackage{makecell}
\usepackage{listings}
\usepackage{algorithm}
\usepackage{algorithmic}
\usepackage{booktabs}
\usepackage{graphicx}
\usepackage[table]{xcolor}
\usepackage{pifont}
\usepackage{xspace}
\usepackage{array}
\usepackage{colortbl}
\usepackage{tabularx}
\usepackage{listings}
\usepackage{amssymb}
\usepackage{enumitem}
\usepackage{colortbl}
\usepackage{xcolor}
\usepackage{booktabs}
\usepackage{array}
\usepackage{pifont}
\usepackage[most]{tcolorbox}
\newcommand{\huggingface}{\raisebox{-1.5pt}{\includegraphics[height=1.05em]{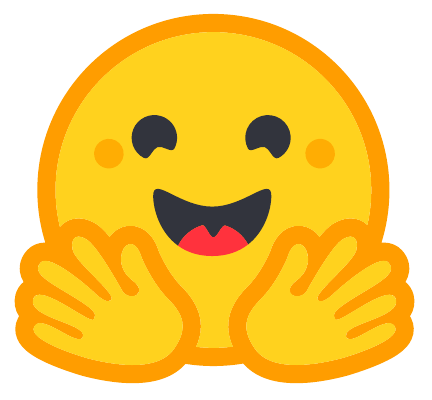}}\xspace}
\newcommand{\github}{\raisebox{-1.5pt}{\includegraphics[height=1.05em]{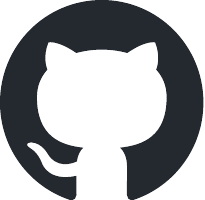}}\xspace}
\definecolor{pyrag-prompt-sys}{HTML}{2E5C8A}
\definecolor{pyrag-prompt-sys-bg}{HTML}{F0F5FB}
\definecolor{pyrag-prompt-user}{HTML}{2F7A4D}
\definecolor{pyrag-prompt-user-bg}{HTML}{F1F8F3}
\definecolor{pyrag-prompt-fix}{HTML}{B4651E}
\definecolor{pyrag-prompt-fix-bg}{HTML}{FBF4EB}
\definecolor{pyrag-placeholder}{HTML}{8E3A8E}

\lstdefinestyle{pyragprompt}{
  basicstyle=\ttfamily\footnotesize,
  breaklines=true,
  breakatwhitespace=true,
  columns=fullflexible,
  keepspaces=true,
  showstringspaces=false,
  upquote=true,
  aboveskip=2pt, belowskip=2pt,
  xleftmargin=2pt, xrightmargin=2pt,
  literate=
  {\{query\}}{{\textcolor{pyrag-placeholder}{\{query\}}}}{7}
  {\{original_query\}}{{\textcolor{pyrag-placeholder}{\{original\_query\}}}}{16}
  {\{sub_queries\}}{{\textcolor{pyrag-placeholder}{\{sub\_queries\}}}}{13}
  {CODE_EXAMPLE}{{\textcolor{pyrag-placeholder}{CODE\_EXAMPLE}}}{12}
  {\{failed_code\}}{{\textcolor{pyrag-placeholder}{\{failed\_code\}}}}{13}
  {\{error_msg\}}{{\textcolor{pyrag-placeholder}{\{error\_msg\}}}}{11}
  {\{error_detail\}}{{\textcolor{pyrag-placeholder}{\{error\_detail\}}}}{14},
}

\definecolor{darkbrown}{RGB}{87, 36, 40}
\definecolor{darkred}{RGB}{176, 46, 54}
\definecolor{darkgreen}{RGB}{42, 102, 6}

\newcommand{\cmark}{\textcolor{green!60!black}{\ding{51}}} 
\newcommand{\xmark}{\textcolor{red!70!black}{\ding{55}}}   
\newcommand{\pmark}{\textcolor{orange!80!black}{$\triangle$}} 

\definecolor{pyblue}{RGB}{53,114,165}
\definecolor{pyyellow}{RGB}{255,184,28}
\newcommand{\pyrag}{{\ttfamily\bfseries\color{pyblue}Py}{\bfseries\color{pyyellow}RAG}\xspace}
\newcommand{\todo}[1]{}

\newcommand{\affillogo}[1]{\textsuperscript{\raisebox{-0.25ex}{\includegraphics[height=0.85em]{#1}}}}
\newcommand{\affillegendlogo}[1]{\raisebox{-0.2ex}{\includegraphics[height=0.95em]{#1}}}
\DeclareRobustCommand{\uiuclogo}{\affillogo{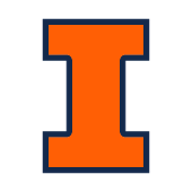}}
\DeclareRobustCommand{\hkustlogo}{\affillogo{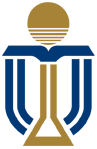}}
\DeclareRobustCommand{\tamulogo}{\affillogo{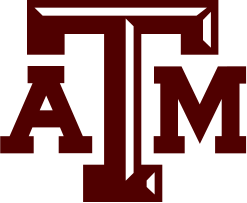}}
\DeclareRobustCommand{\uiuclogoline}{\affillegendlogo{figures/uiuc_logo.png}}
\DeclareRobustCommand{\hkustlogoline}{\affillegendlogo{figures/hkust_logo.png}}
\DeclareRobustCommand{\tamulogoline}{\affillegendlogo{figures/tamu_logo.png}}

\definecolor{pyrag-good}{RGB}{34,139,34}   
\definecolor{pyrag-bad}{RGB}{178,34,34}     
\definecolor{pyrag-step}{RGB}{70,130,180}   
\definecolor{pyrag-bg-good}{RGB}{240,250,240}
\definecolor{pyrag-bg-bad}{RGB}{253,242,242}
\definecolor{pyrag-var}{RGB}{139,69,19}

\lstdefinestyle{pyragcode}{%
  language=Python,
  basicstyle=\ttfamily\footnotesize,
  columns=fullflexible,
  keepspaces=true,
  showstringspaces=false,
  breaklines=true,
  breakatwhitespace=false,
  frame=single,
  framerule=0.4pt,
  rulecolor=\color{black!20},
  backgroundcolor=\color{pyrag-bg-good},
  commentstyle=\color{darkgreen},
  keywordstyle=\color{pyblue}\bfseries,
  stringstyle=\color{darkred},
  xleftmargin=0.5em,
  framexleftmargin=0.5em
}

\lstdefinestyle{pyragcode_warn}{%
  style=pyragcode,
  backgroundcolor=\color{pyrag-bg-bad},
  rulecolor=\color{pyrag-bad!40}
}

\lstdefinestyle{pyragcode_ok}{%
  style=pyragcode,
  backgroundcolor=\color{yellow!15},
  rulecolor=\color{yellow!50!black}
}

\newcommand{\equalcontrib}{\textsuperscript{*}}
\title{Retrieval is Cheap, Show Me the Code: Executable Multi-Hop Reasoning for Retrieval-Augmented Generation}

\author{%
  \textbf{Jiashuo Sun}\equalcontrib\uiuclogo\quad
  \textbf{Jimeng Shi}\equalcontrib\uiuclogo\quad
  \textbf{Yixuan Xie}\uiuclogo\quad
  \textbf{Saizhuo Wang}\hkustlogo\quad
  \textbf{Jash Rajesh Parekh}\uiuclogo \\[3pt]
  \textbf{Pengcheng Jiang}\uiuclogo\quad
  \textbf{Zhiyi Shi}\uiuclogo\quad
  \textbf{Jiajun Fan}\uiuclogo\quad
  \textbf{Qinglong Zheng}\uiuclogo\quad
  \textbf{Peiran Li}\tamulogo \\[3pt]
  \textbf{Shaowen Wang}\uiuclogo\quad
  \textbf{Ge Liu}\uiuclogo\quad
  \textbf{Jiawei Han}\uiuclogo \\[7pt]
  \normalfont\normalsize
  \uiuclogoline~University of Illinois Urbana-Champaign \\
  \hkustlogoline~Hong Kong University of Science and Technology \\
  \tamulogoline~Texas A\&M University \\[6pt]
  \github\ \href{https://github.com/GasolSun36/PyRAG}{\texttt{GitHub}}
  \hspace{3em}
  \huggingface\ \href{https://gasolsun36.github.io/PyRAG/}{\texttt{Project Page}}
  \hspace{3em}
  \huggingface\ \href{https://huggingface.co/gasolsun/PyRAG-7b}{\texttt{Model}}
}
\makeatletter
\g@addto@macro\@thanks{\footnotetext[1]{Equal contribution.}}
\makeatother

\begin{document}

\maketitle

\begin{abstract}
  \label{sec:abstract}
  Retrieval-Augmented Generation (RAG) has become a standard approach for knowledge-intensive question answering, but existing systems remain brittle on multi-hop questions, where solving the task requires chaining multiple retrieval and reasoning steps. Key challenges are that current methods represent reasoning through free-form natural language, where intermediate states are implicit, retrieval queries can drift from intended entities, and errors are detected by the same model that produces them making self-reflection an unreliable, ungrounded signal.
  We observe that multi-hop question answering is a typical form of step-by-step computation, and that this structured process aligns closely with how code-specialized language models are trained to operate. Motivated by this, we introduce \pyrag, a framework that reformulates multi-hop RAG as program synthesis and execution. Instead of free-form reasoning trajectories, \pyrag represents the reasoning process as an executable Python program over retrieval and QA tools, exposing intermediate states as variables, producing deterministic feedback through execution, and yielding an inspectable trace of the entire reasoning process. This formulation further enables compiler-grounded self-repair and execution-driven adaptive retrieval without any additional training.
  Experiments on five QA benchmarks (PopQA, HotpotQA, 2WikiMultihopQA, MuSiQue, and Bamboogle) show that \pyrag consistently outperforms strong baselines under both training-free and RL-trained settings, with especially large gains on compositional multi-hop datasets. Our code, data and models are publicly available at \url{https://github.com/GasolSun36/PyRAG}.
\end{abstract}

\input{introduction}
\input{tables/ours_vs_baselines}

\begin{figure}[t]
  \centering
  \includegraphics[width=\textwidth]{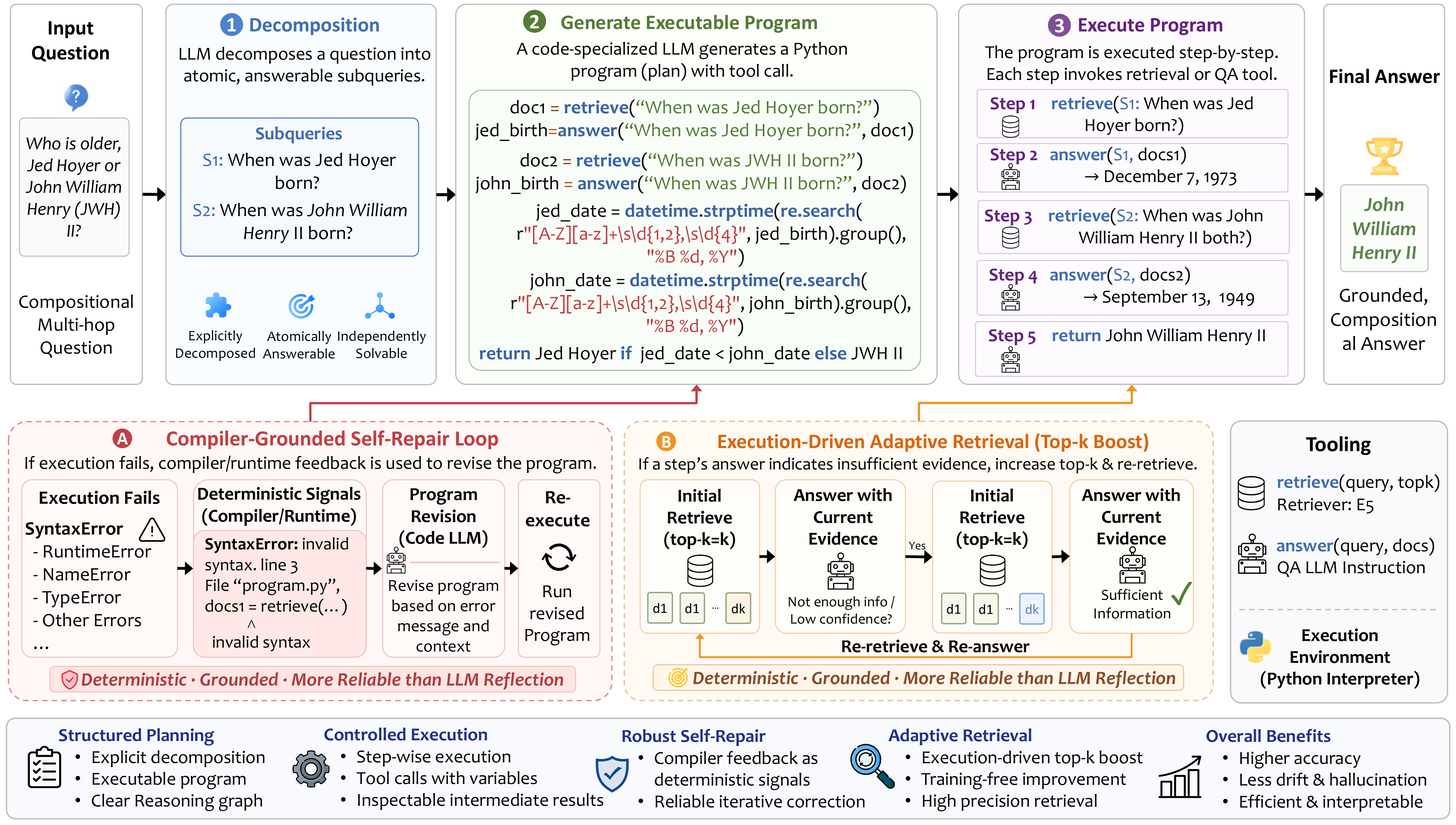}
  \caption{\textbf{The \pyrag framework.}
    Given a multi-hop question, \pyrag proceeds in three stages:
    (1) Decompose: an LLM breaks the question into atomic, independently answerable sub-queries;
    (2) Plan: a code-specialized LLM synthesizes an executable Python program over two tool primitives, \texttt{retrieve(query, topk)} and \texttt{answer(query, docs)}, where intermediate results are bound to variables and composed through explicit data dependencies;
    (3) Execute: the program is run step-by-step in a Python interpreter, producing an inspectable trace and a grounded final answer.
    Two execution-guided refinement mechanisms refine this pipeline:
    (A) Compiler-Grounded Self-Repair, which uses runtime exceptions (e.g., \texttt{SyntaxError}, \texttt{NameError}) as deterministic signals for the planner to revise and re-execute the program; and
  (B) Execution-Driven Adaptive Retrieval, which boosts the top-$k$ retrieval budget for sub-steps whose answer indicates insufficient evidence. Both mechanisms are training-free and rely on grounded execution feedback rather than LLM self-reflection.}
  \label{fig:introduction}
\end{figure}

\input{method}

\input{experiment}

\input{related_work}

\section{Conclusion}
\label{sec:conclusion}

We presented \pyrag, a framework that reformulates multi-hop RAG as program synthesis and execution. By encoding the retrieval--reasoning process as an executable Python program, \pyrag exposes intermediate states as variables, produces deterministic compiler feedback, and yields an inspectable reasoning trace, while enabling training-free self-repair and adaptive retrieval as direct byproducts of the execution interface. Across five QA benchmarks under both training-free and RL-trained settings, \pyrag delivers consistent gains over strong baselines, with the largest improvements on compositional multi-hop datasets.

\bibliographystyle{plainnat}
\bibliography{references}


\appendix

\newpage

\input{limitations}

\section{Extended Related Work}
\label{app:related_work_extended}

\paragraph{Multi-Hop Retrieval-Augmented Generation.}

Multi-hop QA requires chaining evidence across multiple passages, which vanilla RAG ~\cite{lewis2020retrieval} cannot handle in a single retrieval step. Iterative prompting-based methods interleave retrieval with chain-of-thought (CoT) reasoning \cite{wei2022chain} and reasoning-action loops \cite{yao2023react} or decomposed sub-questions \cite{press2023measuring, trivedi2023interleaving, jiang2023active, shao2023enhancing, khattab2022demonstrate, shi2025hypercube, shi2026multicube}. A parallel line of graph-based approaches constructs reasoning structures over retrieved content \cite{edge2024local, gutierrez2024hipporag, chen2026pathrag, parekh2025structure, sun2024thinkongraph, sun2026dynamicrag, sun2026grace, wu2025structurer1dynamicallyleveragingstructural, sun2026tasr, sun2026rethinkingrerankerboundaryawareevidence}. More recent work involves training the search policy with reinforcement learning, optimizing the multi-turn retrieval process end-to-end. Search-R1 \cite{jin2025searchr1} extends DeepSeek-R1 \cite{guo2025deepseek} style training to retrieval with outcome-based rewards and retrieved token masking to stabilize multi-turn updates, and R1-Searcher \cite{song2025r1searcher} similarly incentivizes search invocation through outcome-based RL. StepSearch \cite{zheng2025stepsearch} densifies the RL signal via hop-wise rewards and redundancy penalties. Training-free agentic variants, such as Search-o1 \cite{li2025search}, embed retrieval inside o1-style long CoT, distilling retrieved documents before reinjecting them into the reasoning chain. Across these methods, the retrieval-reasoning interaction remains an implicit trajectory shaped by prompts or rewards, and error detection relies on LLM-generated signals, rather than external verification. \pyrag instead represents the full pipeline as an executable program, making the reasoning structure explicit, dynamic, and verifiable via compiler feedback.

\subsection{Program-Guided Reasoning}

Executable code has proven to be effective for reasoning tasks with well-defined symbolic structure. PAL~\cite{gao2023pal} and Program-of-Thoughts~\cite{chen2022program} offload numerical reasoning to a Python interpreter, separating planning from deterministic execution. Binder~\cite{cheng2022binding} extends this to table QA via unified natural-language and SQL commands, Faithful-CoT~\cite{lyu2023faithful} translates questions into symbolic programs that are then executed by an external solver, and Logic-LM~\cite{pan2023logic} couples an LLM front-end with symbolic solvers for logical reasoning. ProgramFC~\cite{pan2023fact} compiles natural-language claims into Python-style verification, which are then executed by fact-checking modules. These approaches assume that the evidence required for reasoning is available a priori, grounded in self-contained inputs such as tables or closed evidence corpora. A complementary line of work, exemplified by DSPy~\cite{khattab2024dspy}, treats LM pipelines as compilable programs and automatically optimizes their prompts and demonstrations via bootstrapped traces; its HotPotQA case study uses a hand-designed two-hop module with fixed structure. DSPy operates at the level of \textit{pipeline construction and prompt optimization}, whereas \pyrag prescribes a specific \textit{reasoning representation}—a dynamically generated executable program per query—with execution-grounded self-repair and adaptive retrieval as runtime mechanisms; the two are in principle composable. We do not include DSPy as a direct empirical baseline because the two systems target different layers of the stack and rely on incompatible optimization regimes. DSPy's published HotpotQA pipeline is a hand-designed two-hop module with fixed structure, which cannot adapt to questions of varying hop counts (e.g., MuSiQue's mixture of 2–4 hop queries) without manual redesign. More fundamentally, DSPy's strength comes from teleprompter-based prompt and demonstration bootstrapping over a labeled training set—disabling this optimization reduces DSPy to a standard ReAct prompt, while enabling it makes the comparison incommensurable with our training-free setting and operates at a different granularity (prompt-level optimization) than our RL-trained variant (policy-level optimization). We instead view PyRAG and DSPy as composable: PyRAG prescribes the per-query reasoning representation, while DSPy could in principle optimize the prompts of PyRAG's individual agents. We leave this integration to future work.  \pyrag targets a fundamentally different setting, open-domain multi-hop QA, where intermediate answers are unknown at synthesis time and must be dynamically retrieved during execution, with later retrieval queries depending on the results of earlier ones.

\section{Algorithm}
\label{sec:algorithm}

\begin{algorithm}[htbp]
  \caption{\pyrag}
  \label{alg:pyrag}
  \begin{algorithmic}[1]
    \REQUIRE Question $q$; retriever $\mathcal{R}$; tool APIs $\texttt{retrieve}(\cdot,k)$, $\texttt{answer}(\cdot,\cdot)$;
    agents $\mathcal{A}_{\text{dec}}, \mathcal{A}_{\text{plan}}, \mathcal{A}_{\text{ans}}$;
    default top-$k$ $k_0$, boosted top-$k$ $k_1$ ($k_1 > k_0$);
    max repair rounds $T$; sentinel set $\mathcal{S} = \{\text{``unknown''}, \text{``cannot answer''}, \dots\}$
    \ENSURE Final answer $\hat{a}$; execution trace $\tau$

    \STATE \textbf{// Stage 1: Decomposition}
    \STATE $\mathbf{s} = [s_1, \ldots, s_n] \leftarrow \mathcal{A}_{\text{dec}}(q)$ \hfill $\triangleright$ atomic sub-queries

    \STATE \textbf{// Stage 2: Program Synthesis}
    \STATE $\pi \leftarrow \mathcal{A}_{\text{plan}}(q, \mathbf{s})$ \hfill $\triangleright$ executable Python program over \{retrieve, answer\}

    \STATE \textbf{// Stage 3: Execution with grounded refinement}
    \STATE $t \leftarrow 0$, \; $\tau \leftarrow \emptyset$, \; $\text{env} \leftarrow \texttt{PythonInterpreter}()$
    \WHILE{$t \leq T$}
    \STATE $\hat{a}, \tau, \text{err} \leftarrow \textsc{Execute}(\pi, \text{env}, \mathcal{R}, \mathcal{A}_{\text{ans}}, k_0, k_1, \mathcal{S})$
    \IF{$\text{err} = \texttt{None}$}
    \STATE \textbf{break} \hfill $\triangleright$ program executed successfully
    \ELSE
    \STATE $\pi \leftarrow \mathcal{A}_{\text{plan}}(q, \mathbf{s}, \pi, \text{err})$ \hfill $\triangleright$ (A) compiler-grounded self-repair
    \STATE $t \leftarrow t + 1$
    \ENDIF
    \ENDWHILE
    \RETURN $\hat{a}, \tau$

    \vspace{4pt}
    \STATE \textbf{Procedure} \textsc{Execute}($\pi$, env, $\mathcal{R}$, $\mathcal{A}_{\text{ans}}$, $k_0$, $k_1$, $\mathcal{S}$):
    \STATE \quad \textbf{try:}
    \STATE \quad\quad \textbf{for} each statement $\ell \in \pi$ \textbf{do}
    \STATE \quad\quad\quad \textbf{if} $\ell$ is $v \leftarrow \texttt{retrieve}(query, k)$ \textbf{then}
    \STATE \quad\quad\quad\quad $\text{env}[v] \leftarrow \mathcal{R}(query, k)$
    \STATE \quad\quad\quad \textbf{elseif} $\ell$ is $v \leftarrow \texttt{answer}(query, docs)$ \textbf{then}
    \STATE \quad\quad\quad\quad $a \leftarrow \mathcal{A}_{\text{ans}}(query, docs)$
    \STATE \quad\quad\quad\quad \textbf{if} $a \in \mathcal{S}$ \textbf{then} \hfill $\triangleright$ (B) execution-driven adaptive retrieval
    \STATE \quad\quad\quad\quad\quad $docs' \leftarrow \mathcal{R}(query, k_1)$ \hfill $\triangleright$ boost top-$k$ for under-evidenced step
    \STATE \quad\quad\quad\quad\quad $a \leftarrow \mathcal{A}_{\text{ans}}(query, docs')$
    \STATE \quad\quad\quad\quad $\text{env}[v] \leftarrow a$
    \STATE \quad\quad\quad \textbf{else} \hfill $\triangleright$ native Python ops (regex, arithmetic, control flow)
    \STATE \quad\quad\quad\quad evaluate $\ell$ in env
    \STATE \quad\quad\quad append $(\ell, \text{env}[v])$ to $\tau$
    \STATE \quad\quad \textbf{return} $\text{env}[\text{final}], \tau, \texttt{None}$
    \STATE \quad \textbf{except} Exception $e$: \textbf{return} $\texttt{None}, \tau, e$
  \end{algorithmic}
\end{algorithm}

\section{Additional Experiment}
\label{sec:additional_experiment}

\subsection{Implement Details}
\label{sec: implement_details}
\pyrag is implemented as a three-agent pipeline:
a \textbf{Decompose Agent} that breaks the input question into atomic
sub-queries (JSON list, with up to 3 self-correction retries);
a \textbf{Plan Agent} that translates the sub-queries into executable
Python code using two primitive functions, \texttt{retrieve()} and
\texttt{answer()}; and an \textbf{Answer Agent} that processes each
\texttt{answer()} call by conditioning on retrieved passages enclosed
in structured \texttt{<answer>} tags.
All inter-agent communication is mediated through a shared
\texttt{execution\_log} that records every retrieval and QA step.

Runtime errors in LLM-generated code trigger a self-repair loop in
which the Plan Agent is re-prompted with the failed code and the
Python traceback, up to $\texttt{MAX\_FIX\_ROUNDS}{=}3$ attempts.
Syntax errors detected during code generation are corrected inline
within the same generation call, also up to 3 retries.

All language models are served with vLLM~\citep{kwon2025vllm}.
The Plan Agent uses Qwen2.5-Coder-7B-Instruct~\citep{hui2024qwen25coder} (tensor parallel
size 2); the Decompose and Answer Agents use Qwen2.5-7B-Instruct~\cite{qwen2025qwen25}
(tensor parallel size 2).
For 72B-backbone experiments, Qwen2.5-72B-Instruct is substituted
with tensor parallel size 4.

We fine-tune all three agents with GRPO~\citep{shao2024deepseekmath} using the VERL framework~\citep{sheng2025hybridflow} under a shared-parameter, curriculum-style schedule: a single backbone is sequentially specialized into the Answer, Plan, and Decompose roles, with the other two agents frozen at each stage. The order is deliberate: the Answer Agent is trained first since it is the terminal step of every reasoning chain and its quality bounds the end-to-end reward; the Plan Agent is trained next on top of a well-calibrated answerer, so program-level credit assignment is conditioned on a reliable execution backend; finally, the Decompose Agent is trained against frozen Plan and Answer Agents that are both already strong, which substantially reduces the variance of the end-to-end reward signal. The reward is a weighted combination of EM and F1, $r = 0.7 \cdot \mathrm{F1} + 0.3 \cdot \mathrm{EM}$, computed by executing the full pipeline against gold answers. All three agents are fine-tuned with LoRA~\citep{hu2022lora} (rank 64, $\alpha=32$). The Answer Agent uses learning rate $1\mathrm{e}{-6}$, cosine schedule, rollout $n=8$, batch size $32$, $1$ epoch. The Plan and Decompose Agents use batch size $64$, rollout $n=4$, learning rate $3\mathrm{e}{-6}$, KL penalty $\lambda=0.001$ (low-variance KL), $2$ epochs each. All RL experiments are conducted on a single node of $8\times$A100 80\,GB GPUs.

\subsection{Datasets}
Following the data setup of Search-R1~\citep{jin2025searchr1}, we train on a
mixture of Natural Questions (NQ)~\citep{kwiatkowski2019nq} and
HotpotQA~\citep{yang2018hotpotqa}, yielding 87{,}925 training examples in total
(79{,}168 single-hop NQ and 8{,}757 multi-hop HotpotQA). This mixture exposes the
model to both single-hop factoid retrieval and compositional multi-hop reasoning
during RL fine-tuning, while keeping the training distribution comparable to
prior RL-trained RAG baselines for fair comparison.

For evaluation, we use seven datasets grouped along two axes domain
(in- vs.\ out-of-domain relative to training) and hop count (single- vs.\
multi-hop). The in-domain evaluation sets are the held-out splits of NQ (3{,}610)
and HotpotQA (7{,}405). For out-of-domain evaluation, we include three single-hop
benchmarks: PopQA~\citep{mallen2023popqa} (14{,}267), and three multi-hop benchmarks,
2WikiMultiHopQA~\citep{ho2020twowiki} (12{,}576),
MuSiQue~\citep{trivedi2022musique} (2{,}417), and Bamboogle~\citep{press2023measuring}
(125). The multi-hop out-of-domain sets in particular stress-test whether the
structured planning prior learned by PyRAG transfers across question
distributions, hop counts (MuSiQue contains 2--4 hop questions), and
compositional patterns unseen during training. Detailed statistics are
reported in Table~\ref{tab:dataset_stats}. We use Exact Match (EM) as the
primary metric throughout, consistent with prior work.

\begin{table}[t]
  \centering
  \caption{Dataset statistics for training and evaluation.}
  \label{tab:dataset_stats}
  \begin{tabular}{llrll}
    \toprule
    \textbf{Split} & \textbf{Dataset} & \textbf{\#Examples} & \textbf{Task Type} & \textbf{Domain} \\
    \midrule
    \multirow{3}{*}{Train}
    & NQ          & 79,168  & Single-hop & In-domain \\
    & HotpotQA    & 8,757   & Multi-hop  & In-domain \\
    & \textit{Total} & \textit{87,925} & --        & --        \\
    \midrule
    \multirow{7}{*}{Eval}
    & HotpotQA        & 7,405   & Multi-hop  & In-domain     \\
    \cmidrule(lr){2-5}
    & PopQA           & 14,267  & Single-hop & Out-of-domain \\
    & 2WikiMultiHopQA & 12,576  & Multi-hop  & Out-of-domain \\
    & Musique         & 2,417   & Multi-hop  & Out-of-domain \\
    & Bamboogle       & 125     & Multi-hop  & Out-of-domain \\
    \bottomrule
  \end{tabular}
\end{table}

\section{Prompts}
\label{sec: prompts}

\input{tables/prompt}

\section{Case Study}
\label{sec: case_study}

\input{tables/case_study}



\end{document}

%% file: introduction.tex
\section{Introduction}
\label{sec:introduction}
\begin{figure}[t]
  \centering
  \includegraphics[width=\textwidth]{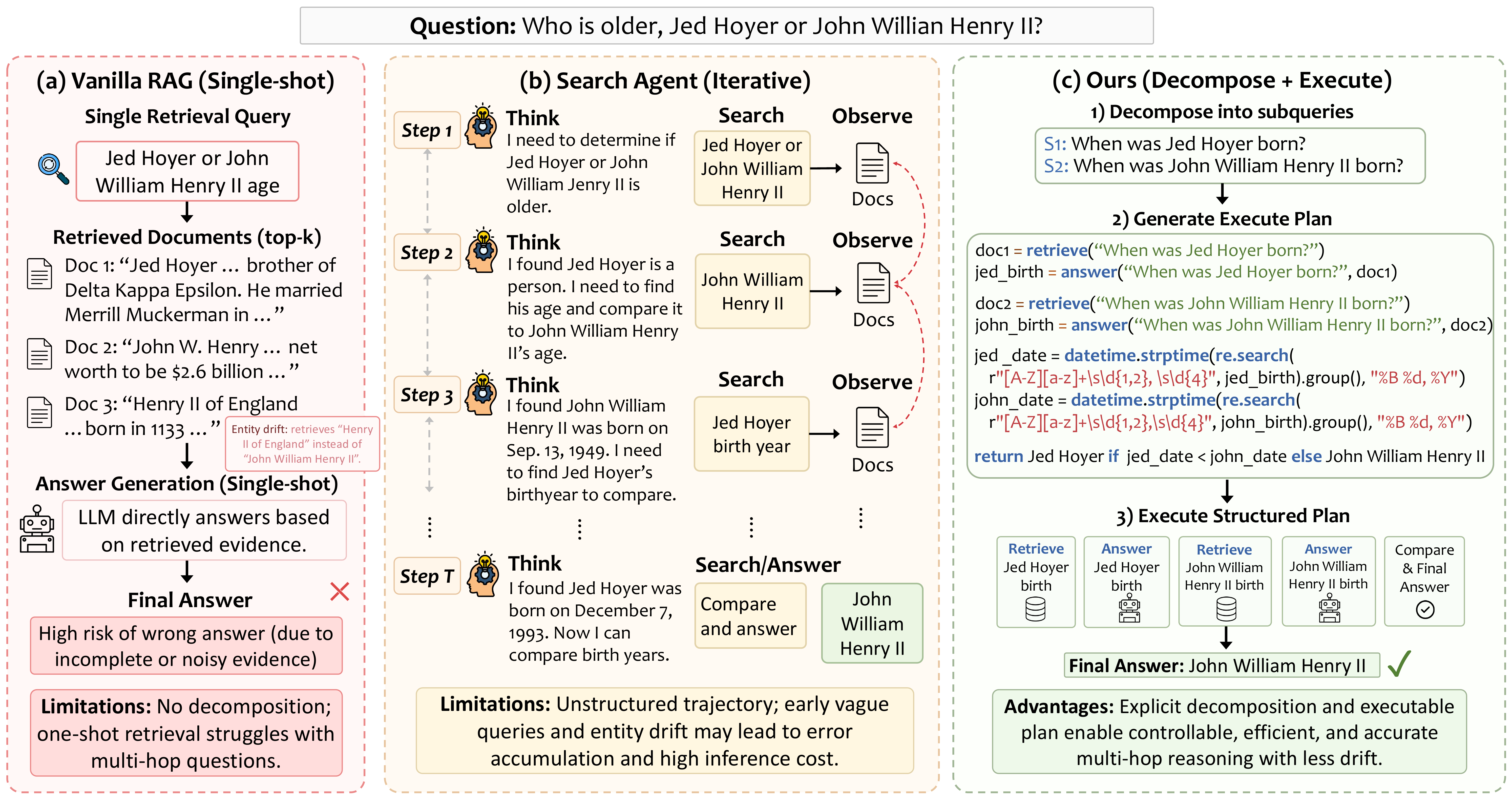}
  \vspace{-4mm}
  \caption{\textbf{Comparison across Vanilla RAG, Search Agents, and \pyrag (Ours).}
    Given the multi-hop question \textit{``Who is older, Jed Hoyer or John William Henry II?''},
    (a) Vanilla RAG performs single-shot retrieval and is prone to incomplete or noisy evidence;
    (b) Search Agents follow an unstructured iterative trajectory where vague queries and entity drift (e.g., retrieving ``Henry II of England'' instead of ``John William Henry II'') accumulate errors across steps;
  (c) \pyrag decomposes the question into atomic sub-queries and generates an executable program that retrieves, answers, and composes intermediate results through explicit variables, yielding a controllable, inspectable, and accurate reasoning process.}
  \label{fig:comparison}
\end{figure}

Retrieval-Augmented Generation (RAG)~\citep{gao2023retrieval, lewis2020retrieval} has emerged as a foundational paradigm for knowledge-intensive question answering, allowing large language models (LLMs) to ground their outputs in external evidence and produce more factual responses~\citep{fan2024survey, hendrycks2020measuring}. While vanilla RAG works well for single-hop queries, many real-world questions require multi-hop reasoning~\citep{ho2020twowiki, yang2018hotpotqa, trivedi2022musique, press2023measuring, shi2026multicube}, where the answer must be assembled by chaining evidence across multiple sources. For example, answering \textit{``Who is older, Jed Hoyer or John William Henry II?''} requires retrieving two birth dates, maintaining them as intermediate results, and composing them through an explicit comparison. Such questions are pervasive in open-domain QA and stress-test a system's ability to plan, retrieve iteratively, and aggregate evidence across steps. Figure~\ref{fig:comparison} illustrates how three representative paradigms, Vanilla RAG, Search Agents, and our \pyrag, approach this question, highlighting the structural differences in how each maintains intermediate state and composes evidence.

Existing multi-hop RAG approaches are typically achieved via free-form natural language reasoning, including chain-of-thought prompting~\citep{wei2022chain}, iterative retrieve-and-reason loops~\citep{trivedi2023interleaving, yao2023react, shao2023enhancing, shi2026multicube}, and more recently, reinforcement-learned search agents~\citep{jin2025searchr1, song2025r1searcher, zheng2025stepsearch, chen2025research}. While these methods introduce decomposition and iteration, the reasoning state remains implicit in text: intermediate results are embedded in narrative form rather than maintained as discrete objects, retrieval queries can drift from the intended entities (e.g., querying \textit{``Henry II of England''} when the question concerns \textit{``John William Henry II''}), and errors are detected by the same LLM that produces them, turning self-reflection into an unreliable, ungrounded signal. As a result, the reasoning trajectory is hard to control, verify, and troubleshoot.
Although a parallel line of program-guided reasoning work~\citep{gao2023pal, chen2022program, cheng2022binding, lyu2023faithful, pan2023logic, pan2023fact} does leverage executable code, they assume that the evidence required for reasoning is available a priori in self-contained inputs such as tables or closed corpora.
This assumption breaks down in open-domain multi-hop QA, where intermediate answers are unknown at synthesis time, and subsequent queries must depend on the results of earlier retrievals.
Table~\ref{tab:paradigm_comparison1} summarizes how these reasoning paradigms differ along five key dimensions: multi-hop capability, interpretability, structured planning, reflection, and executable interface, motivating the design of \pyrag as a paradigm that supports all five.

We argue that the root cause of these limitations is a mismatch between task structure and reasoning representation. Multi-hop question answering is fundamentally a form of step-by-step computation: it decomposes a question into sub-problems, computes intermediate results, and composes them through explicit dependencies.
This process mirrors how programs are constructed and executed, a sequence of operations over named variables, connected by data flow.
Yet, current methods simply encode this structured computation into unstructured natural language by forcing the LLM to simultaneously plan, maintain state, and reason.
We further observe that code-specialized language models are explicitly trained for this exact pattern of behavior: maintaining intermediate variables, enforcing control flow, and producing step-by-step structured programs~\citep{huang2024code}. This suggests a natural reformulation: if we represent multi-hop reasoning as program synthesis rather than free-form generation, we can directly leverage the inductive bias of code models, while simultaneously gaining explicit state, deterministic feedback from execution, and an inspectable trace of the reasoning process.

Motivated by this observation, we introduce \pyrag, a framework that provides a verifiable execution interface for multi-hop RAG. \pyrag casts multi-hop reasoning as the synthesis and execution of a Python program over a small set of tool APIs: \texttt{retrieve(query)} and \texttt{answer(query, docs)}, where each step retrieves evidence, computes an intermediate answer, and stores the result as a variable that can be reused downstream. The framework consists of three specialized agents: a \texttt{Decompose Agent} that breaks the input question into atomic sub-queries, a \texttt{Plan Agent} that translates the sub-queries into an executable program, and an \texttt{Answer Agent} that produces short answers from retrieved evidence. Crucially, the executable formulation gives rise to two natural refinement mechanisms with no additional training: a \textit{compiler-grounded self-repair loop}, where runtime exceptions provide deterministic signals for the Plan Agent to revise the program, and an \textit{execution-driven adaptive retrieval} mechanism that selectively increases the retrieval scope when an intermediate answer indicates insufficient evidence. Both arise directly from the program-execution interface rather than relying on LLM self-reflection.

We evaluate \pyrag on five open-domain QA benchmarks (PopQA, HotpotQA, 2WikiMultihopQA, MuSiQue, Bamboogle) under both training-free and RL-trained settings. Our contributions are:
\begin{itemize}[leftmargin=*, itemsep=2pt, topsep=2pt]
  \item We identify a structural mismatch between multi-hop reasoning and its representation in existing RAG systems, and reformulate multi-hop QA as an \textbf{executable} step-by-step process.
  \item We introduce \pyrag, a framework that provides a verifiable execution interface with explicit state, deterministic compiler feedback, and inspectable reasoning traces, equipped with execution-guided self-repair and adaptive retrieval.
  \item We show that the advantage of code-specialized models is task-dependent: it emerges only under program-synthesis interfaces, highlighting that model capability and reasoning interface must be co-designed.
  \item Empirically, \pyrag improves over Vanilla RAG by +11.8 average EM (training-free, 7B) and +25.5 on Bamboogle, while \pyrag-RL achieves the highest average EM among 7B-scale RL-trained methods and generalizes across Qwen3-4B and LLaMA-3.1-8B backbones.
\end{itemize}

%% file: tables/ours_vs_baselines.tex
\begin{table*}[t]
\centering
\caption{\textbf{Comparison of reasoning paradigms.} 
Search Agents are marked partial on Structured Planning because their plans are implicit and unfold reactively in the thought trace rather than being materialized as an explicit, inspectable artifact, and partial on Reflection because error signals come from the LLM's own self-judgment rather than grounded external feedback. They lack an Executable Interface entirely, as the reasoning trajectory is natural language with no variables, data flow, or deterministic execution. Our \pyrag addresses all three limitations by representing the multi-hop reasoning process as an executable program.}
\setlength{\tabcolsep}{8pt}
\renewcommand{\arraystretch}{1.4}
\resizebox{\textwidth}{!}{
\begin{tabular}{lccccc}
\toprule
\rowcolor{gray!10}
\textbf{Paradigm} 
& \textbf{Multi-hop Capability} 
& \textbf{Interpretability}
& \textbf{Structured Planning} 
& \textbf{Reflection}
&  \textbf{Executable Interface} \\
\midrule

Vanilla RAG (Single-shot)
& \xmark 
& \xmark 
& \xmark 
& \xmark 
& \xmark \\

Search Agent (Free-form)
& \cmark 
& \cmark 
& \pmark 
& \pmark 
& \xmark \\

\pyrag (Executable program)
& \cmark 
& \cmark 
& \cmark 
& \cmark 
& \cmark \\
\bottomrule
\end{tabular}
}
\vspace{2pt}
\footnotesize{\cmark: supported \quad \pmark: partially supported \quad \xmark: not supported}
\label{tab:paradigm_comparison1}
\end{table*}

%% file: method.tex
\section{Method}
\label{sec:method}
\subsection{Overview}

We present \pyrag, a framework that introduces an executable interface for multi-hop RAG, as shown in Figure \ref{fig:introduction}.
Instead of representing reasoning as free-form natural language, \pyrag decomposes the problem into
a sequence of structured steps and executes them through a program.

Given a question $q$, \pyrag consists of three components:
(1) a \textbf{decomposition agent} that breaks $q$ into atomic sub-queries,
(2) a \textbf{planning agent} that generates an executable program describing the reasoning process,
and (3) an \textbf{answer agent} that produces answers based on retrieved evidence.

At the inference time, the generated program is executed step-by-step, where each step corresponds to
a retrieval or question-answering operation.
Therefore, we shift multi-hop reasoning from opaque narrative to an explicit, controllable, and verifiable execution process.

\subsection{Motivation: Multi-Hop QA as Step-by-Step Computation}

We argue that multi-hop question answering can be naturally viewed as a form of
step-by-step computation.
Resolving multi-hop queries necessitates a systematic decomposition into constituent sub-problems, the computation of intermediate results, and the ultimate synthesis of these findings into a final answer~\citep{trivedi2023interleaving, yao2023react, shao2023enhancing, wei2022chain}.

This structured process closely aligns with the fundamental principles of programmatic execution:
A program defines a sequence of functional operations, maintains intermediate variables, and enforces dependencies between steps~\citep{huang2024code}.
Code-specialized language models are explicitly trained for such behavior.
They are optimized to generate structured programs that decompose tasks,
maintain state through variables, and perform consistent step-by-step execution.
As a result, they provide a strong inductive bias for multi-hop QA processing.

Motivated by this observation, we cast multi-hop RAG as a program synthesis problem,
where the reasoning process is represented as an executable plan. This allows us to
directly leverage the step-by-step reasoning capability of code models for explicit control and verification, rather than
forcing it to emerge from free-form natural language reasoning.

\subsection{\pyrag Agents}

\paragraph{Decomposition Agent.}
Given a question $q$, the decomposition agent produces a sequence of sub-queries
$s = [s_1, \ldots, s_n]$, where each sub-query is designed to be answerable with a single retrieval step.
This step introduces an explicit structure over the reasoning process, but does not yet define how
the steps should be executed or combined.

\paragraph{Answer Agent.}
The answer agent takes a sub-query and a set of retrieved documents as input, and produces a short answer.
It is implemented using an instruction-following LLM, and is responsible for extracting information
from retrieved evidence and performing final aggregation.

\paragraph{Planning Agent.}
The planning agent is the core component of \pyrag. Given the original question $q$ and the
decomposed sub-queries $s$, it generates a program $\pi$ that specifies how to solve the task
through a sequence of retrieval and answering operations.

\subsection{Executable Planning}

We define two APIs for the execution tool:
\begin{itemize}[leftmargin=*, itemsep=2pt, topsep=2pt]
  \item \texttt{retrieve(query, topk=k)}: returns the top-k relevant documents for the given query, where k can be increased adaptively at execution time (Sec. \ref{sec:adaptive_retrieval}).
  \item \texttt{answer(query, docs)}: returns an answer conditioned on documents.
\end{itemize}

The planning agent generates a program that composes these APIs through variable assignments.
Each step retrieves evidence, computes an intermediate answer, and stores the result in a variable.
These variables are then reused in subsequent steps.

This formulation makes the reasoning process explicit:
instead of implicitly encoding intermediate states in text, the program stores them as variables
and connects them through data dependencies. The final answer is produced by aggregating these
intermediate results.

\subsection{Execution}

The generated program $\pi$ is executed step-by-step. At each step, the system invokes either
\texttt{retrieve()} or \texttt{answer()}, and stores the output for later use.

This execution process yields an execution trace, which records all intermediate queries,
retrieved documents, and answers. The trace provides a transparent view of the reasoning process
and enables debugging and analysis.

\subsection{Execution-Guided Reflexion}

A key advantage of executable planning is that it naturally supports refinement during execution.

\paragraph{Compiler-Grounded Self-Repair}
If the generated program fails to execute due to invalid operations or inconsistent variable usage,
the execution environment returns a structured error signal. The planning agent can then revise the
program based on this feedback and re-execute it.

\paragraph{Adaptive Retrieval}
\label{sec:adaptive_retrieval}
If an intermediate answer indicates insufficient evidence, the system can selectively increase the
retrieval scope for that step and re-run the corresponding operation. This allows targeted correction
without modifying the entire reasoning plan.

These mechanisms arise naturally from the executable formulation, without requiring additional
training or specialized control logic.

%% file: experiment.tex
\section{Experiments}
\label{sec:experiments}

\subsection{Experimental Setup}

\paragraph{Benchmarks.}
We evaluate on five open-domain QA benchmarks spanning single-hop and
multi-hop reasoning:
PopQA~\citep{mallen2023popqa},
HotpotQA~\citep{yang2018hotpotqa},
2WikiMultihopQA~\citep{ho2020twowiki},
MuSiQue~\citep{trivedi2022musique}, and
Bamboogle~\citep{press2023measuring}.

Exact Match (EM) is used as the primary metric for all benchmarks.
HotpotQA serves as the in-domain training set for RL-trained variants;
all remaining datasets are evaluated out-of-domain.

\paragraph{Baselines.}
We compare against the following categories of methods:

\noindent\textit{Training-free baselines.}
\textbf{Direct Inference} and \textbf{CoT}~\citep{wei2022chain} require
no retrieval.
\textbf{Vanilla RAG}~\citep{lewis2020retrieval} performs single-step retrieve-then-read.
\textbf{Self-Ask}~\citep{press2023measuring} decomposes questions into
sub-questions with interleaved retrieval.
\textbf{IRCoT}~\citep{trivedi2023interleaving} interleaves chain-of-thought
reasoning with iterative retrieval.
\textbf{ITER-RETGEN}~\citep{shao2023enhancing} alternates between retrieval and generation across
multiple rounds.

\noindent\textit{RL-trained baselines.}
\textbf{RAG-SFT} and \textbf{RAG-RL} are supervised fine-tuning and
reinforcement learning variants of a standard RAG pipeline.
\textbf{ZEROSEARCH}~\citep{sun2025zerosearch}, \textbf{Search-R1}~\citep{jin2025searchr1},
\textbf{StepSearch}~\citep{zheng2025stepsearch}, and \textbf{ReSearch}~\citep{chen2025research} are
recent RL-based methods that train models to perform adaptive retrieval.

\noindent\textit{Our methods.}
\textbf{\pyrag} is our training-free multi-agent framework;
\textbf{\pyrag-RL} further fine-tunes the framework with reinforcement
learning.
Unless stated otherwise, all \pyrag variants use Qwen2.5-7B-Instruct
as the backbone.

\paragraph{Implementation Details.}

We follow the retrieval and data setup of Search-R1~\citep{jin2025searchr1}
exactly: an E5-base dense retriever over the Wikipedia 2018 dump~\citep{karpukhin2020dpr},
with the same training splits and evaluation data preprocessing.
The default number of retrieved passages per sub-query is $k{=}5$.
When an \texttt{answer()} call returns an insufficient-information
response (e.g.\ ``unknown'' or ``cannot answer''), the runner
automatically re-executes the same code with an increased retrieval
budget of $k{=}10$ for the implicated steps. Additional implementation details including training are provided in Appendix \ref{sec: implement_details}.

\begin{table}[t]
  \centering
  \setlength{\tabcolsep}{4pt}
  \renewcommand{\arraystretch}{1.1}
  \caption{Exact Match (\%) of \textbf{training-free} methods on five QA benchmarks.
    All methods use the same setting and identical
    evaluation splits. Best result within each backbone size is in
    \textbf{bold}; second best is \underline{underlined}.
  $\Delta$ shows improvement over Vanilla RAG within the same backbone.}
  \resizebox{0.9\textwidth}{!}{%
    \begin{tabular}{lcccccc}
      \toprule
      \textbf{Method} & \textbf{PopQA} & \textbf{HotpotQA} & \textbf{2WikiMQA} & \textbf{MuSiQue} & \textbf{Bamboogle} & \textbf{Avg.} \\
      \midrule
      \multicolumn{7}{c}{\textit{Backbone: Qwen2.5-7B-Instruct}} \\
      \midrule
      Direct Inference             & 14.0 & 18.3 & 12.6 & 3.1  & 12.0 & 12.0 \\
      CoT~\citep{wei2022chain}     & 5.4  & 9.2  & 10.8 & 2.2  & 23.2 & 10.2 \\
      Vanilla RAG                  & 26.7 & 28.9 & 18.9 & 4.7  & 16.0 & 19.0 \\
      Self-Ask~\citep{press2023measuring}      & 29.4 & 30.2 & 21.5 & 6.8  & 22.1 & 22.0 \\
      IRCoT~\citep{trivedi2023interleaving}    & \underline{32.6} & \underline{32.7} & 24.8 & \underline{9.1}  & 24.3 & 24.7 \\
      ITER-RETGEN~\citep{shao2023enhancing}                  & 31.4 & 32.5 & \underline{28.9} & 8.7  & \underline{29.6} & \underline{26.2} \\
      \rowcolor{gray!10}
      \textbf{\pyrag (ours)}        & \textbf{33.5} & \textbf{34.0} & \textbf{33.4} & \textbf{11.8} & \textbf{41.5} & \textbf{30.8} \\
      \rowcolor{gray!10}
      \textit{$\Delta$ vs.\ Vanilla RAG} & \textcolor{green!60!black}{\textbf{+6.8}} & \textcolor{green!60!black}{\textbf{+5.1}} & \textcolor{green!60!black}{\textbf{+14.5}} & \textcolor{green!60!black}{\textbf{+7.1}} & \textcolor{green!60!black}{\textbf{+25.5}} & \textcolor{green!60!black}{\textbf{+11.8}} \\
      \midrule
      \multicolumn{7}{c}{\textit{Backbone: Qwen2.5-72B-Instruct}} \\
      \midrule
      Direct Inference             & 19.7 & 30.6 & 20.6 & 5.5  & 17.6 & 18.8 \\
      CoT~\citep{wei2022chain}     & 24.4 & 33.2 & 25.4 & 9.9  & 19.6 & 22.5 \\
      Vanilla RAG                  & 33.2 & 36.8 & 30.4 & 10.6 & 21.6 & 26.5 \\
      Self-Ask~\citep{press2023measuring}      & 41.4 & 48.2 & 32.5 & 11.8 & 26.1 & 32.0 \\
      IRCoT~\citep{trivedi2023interleaving}    & \underline{44.6} & \underline{50.9} & 35.8 & \underline{14.2} & 28.3 & 34.8 \\
      ITER-RETGEN~\citep{shao2023enhancing}                  & 43.4 & 50.5 & \underline{40.2} & 13.8 & \underline{33.6} & \underline{36.3} \\
      \rowcolor{gray!10}
      \textbf{\pyrag (ours)}        & \textbf{45.5} & \textbf{52.0} & \textbf{44.4} & \textbf{16.9} & \textbf{45.5} & \textbf{40.9} \\
      \rowcolor{gray!10}
      \textit{$\Delta$ vs.\ Vanilla RAG} & \textcolor{green!60!black}{\textbf{+12.3}} & \textcolor{green!60!black}{\textbf{+15.2}} & \textcolor{green!60!black}{\textbf{+14.0}} & \textcolor{green!60!black}{\textbf{+6.3}} & \textcolor{green!60!black}{\textbf{+23.9}} & \textcolor{green!60!black}{\textbf{+14.4}} \\
      \bottomrule
    \end{tabular}%
  }
  \vspace{2mm}

  \label{tab:main_training_free}
\end{table}

\begin{table}[t]
  \centering
  \setlength{\tabcolsep}{4pt}
  \renewcommand{\arraystretch}{1.1}
  \caption{Exact Match (\%) of RL-trained methods on four QA benchmarks.
    All methods are evaluated
    under the same retrieval setting.
    The best result within each backbone is in \textbf{bold}.
  $^\dagger$ denotes in-domain evaluation; remaining datasets are out-of-domain.}
  \resizebox{0.9\textwidth}{!}{%
    \begin{tabular}{lccccc}
      \toprule
      \textbf{Method} & \textbf{HotpotQA$^\dagger$} & \textbf{2WikiMQA} & \textbf{MuSiQue} & \textbf{Bamboogle} & \textbf{Avg.} \\
      \midrule
      \multicolumn{6}{c}{\textit{Backbone: Qwen2.5-7B-Instruct}} \\
      \midrule
      Vanilla RAG                              & 28.9 & 18.9 & 4.7 & 16.0 & 21.3 \\
      RAG-SFT                                 & 32.4 & 22.6 & 6.8 & 27.1 & 22.2 \\
      RAG-RL                                   & 35.2 & 34.7 & 9.6 & 29.6 & 27.3 \\
      ZEROSEARCH~\citep{sun2025zerosearch}                            & 34.6 & 35.2 & 18.4 & 27.7 & 29.0 \\
      Search-R1~\citep{jin2025searchr1}        & 37.0 & 41.4 & 14.6 & 36.8 & 32.4 \\
      StepSearch~\citep{zheng2025stepsearch}                           & 38.6 & 36.6 & \textbf{22.6} & 40.0 & 34.5 \\
      ReSearch~\citep{chen2025research}        & \textbf{43.5} & 47.6 & 22.3 & 42.4 & 38.9 \\
      \rowcolor{gray!10}
      \textbf{\pyrag-RL (ours)}                 & 40.5 & \textbf{49.4} & 20.7 & \textbf{46.1} & \textbf{39.2} \\
      \midrule
      \multicolumn{6}{c}{\textit{Backbone: Qwen3-4B-Instruct}} \\
      \midrule
      Vanilla RAG                              & 27.1 & 16.7 & 4.3  & 14.8 & 15.7 \\
      RAG-SFT                                  & 30.5 & 20.1 & 6.2  & 25.4 & 20.6 \\
      RAG-RL                                   & 33.2 & 31.8 & 8.8  & 27.6 & 25.4 \\
      \rowcolor{gray!10}
      \textbf{\pyrag-RL (ours)}                 & \textbf{38.4} & \textbf{45.1} & \textbf{18.6} & \textbf{43.2} & \textbf{36.3} \\
      \midrule
      \multicolumn{6}{c}{\textit{Backbone: LLaMA-3.1-8B-Instruct}} \\
      \midrule
      Vanilla RAG                              & 30.3 & 19.4 & 6.3  & 17.6 & 18.4 \\
      RAG-SFT                                  & 34.1 & 23.3 & 8.5  & 29.3 & 23.8 \\
      RAG-RL                                   & 37.5 & 35.2 & 11.4 & 31.8 & 29.0 \\
      \rowcolor{gray!10}
      \textbf{\pyrag-RL (ours)}                 & \textbf{43.2} & \textbf{50.1} & \textbf{22.1} & \textbf{48.3} & \textbf{40.9} \\
      \bottomrule
    \end{tabular}%
  }
  \vspace{2mm}

  \label{tab:main_rl}
\end{table}

\subsection{Main Results}
\label{sec:main_results}

Table~\ref{tab:main_training_free} and Table~\ref{tab:main_rl} report
the main results under training-free and RL-trained settings,
respectively.

\paragraph{Training-free results.}
Under the training-free setting (Table~\ref{tab:main_training_free}),
\pyrag consistently outperforms all baselines across both backbone sizes.
With Qwen2.5-7B-Instruct, \pyrag achieves an average EM of 30.8,
surpassing the strongest baseline ITER-RETGEN by +4.6 points
and Vanilla RAG by +11.8 points.
Gains are most pronounced on compositional multi-hop benchmarks:
+14.5 on 2WikiMQA and +25.5 on Bamboogle relative
to Vanilla RAG, datasets specifically designed to stress systems that
cannot chain multiple retrieval steps.
On PopQA and HotpotQA, \pyrag also achieves the best results
(33.5 and 34.0), demonstrating that the structured
decompose-plan-answer pipeline does not degrade performance on
relatively simpler queries.
Scaling to Qwen2.5-72B-Instruct amplifies these trends: \pyrag reaches
an average of 40.9, outperforming ITER-RETGEN by +4.6
and delivering the largest single-dataset gain on Bamboogle
(+23.9 over Vanilla RAG).

\paragraph{RL-trained results.}
Table~\ref{tab:main_rl} compares \pyrag trained with reinforcement
learning (\pyrag-RL) against competitive RL- and SFT-based baselines.
With the Qwen2.5-7B backbone, \pyrag-RL achieves an average EM of
39.2, on par with ReSearch (38.9) while outperforming
all other baselines including Search-R1 (+6.8) and StepSearch (+4.7).
Notably, \pyrag-RL attains the highest score on 2WikiMQA (49.4) and Bamboogle (46.1) among 7B models, while remaining competitive on the HotpotQA and MuSiQue.
\pyrag-RL generalizes well across architectures: it achieves
36.3 average EM on Qwen3-4B and 40.9 on
LLaMA-3.1-8B, consistently surpassing the corresponding RAG-RL
baselines by +10.9 and +11.9 points, respectively,
confirming that the structured planning prior of \pyrag translates
effectively to the RL fine-tuning regime.

\subsection{Ablation Study}
\label{sec:ablation}

\begin{figure}[t]
  \centering

  \begin{subfigure}[t]{0.49\linewidth}
    \centering
    \includegraphics[width=\linewidth]{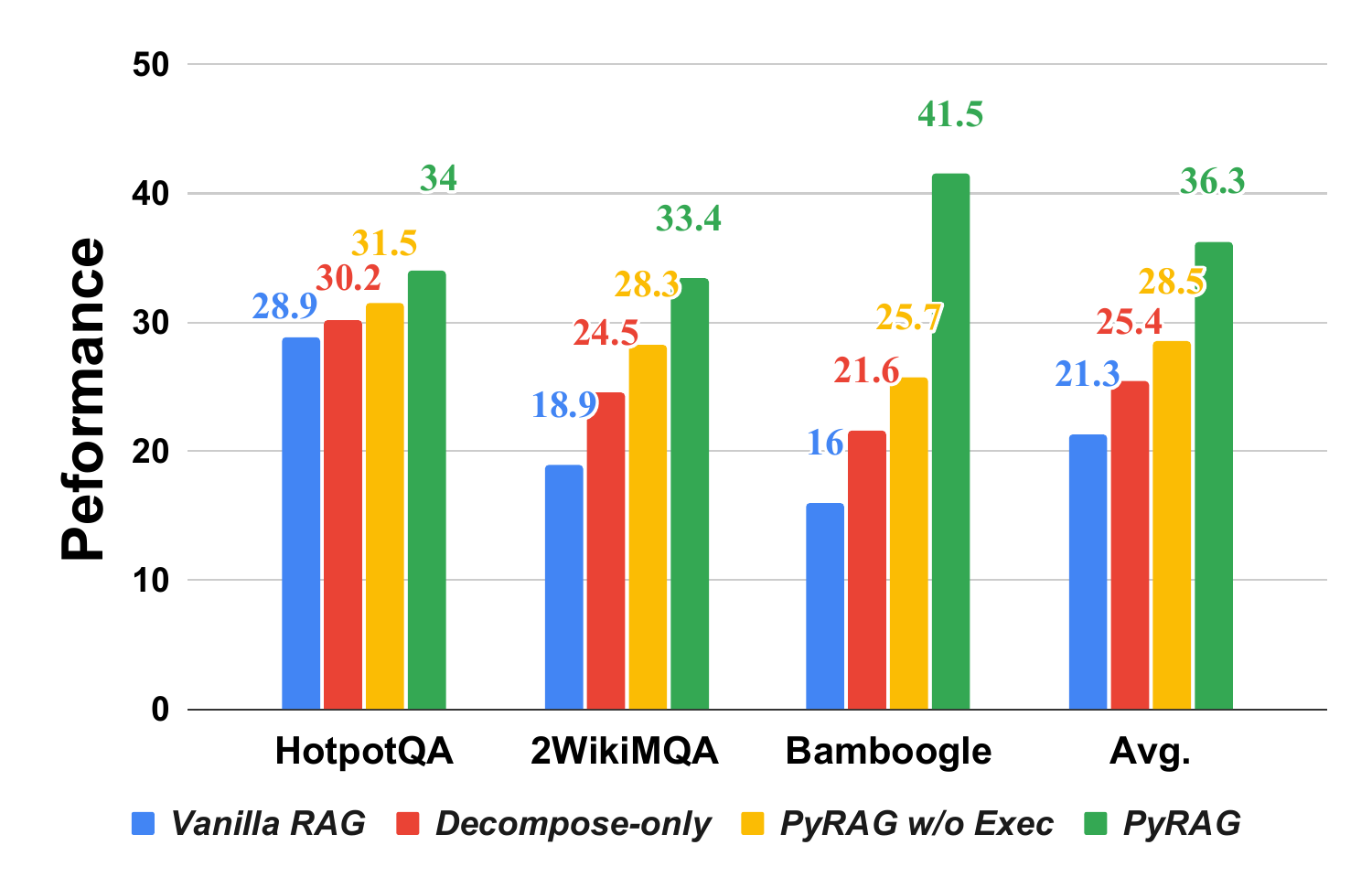}
    \caption{Adding decomposition, planning, and execution to Vanilla RAG yields monotonic gains from 21.3 to 36.3 average EM, with execution contributing the largest jump.}
    \label{fig:progressive_ablation}
  \end{subfigure}
  \hfill
  \begin{subfigure}[t]{0.49\linewidth}
    \centering
    \includegraphics[width=\linewidth]{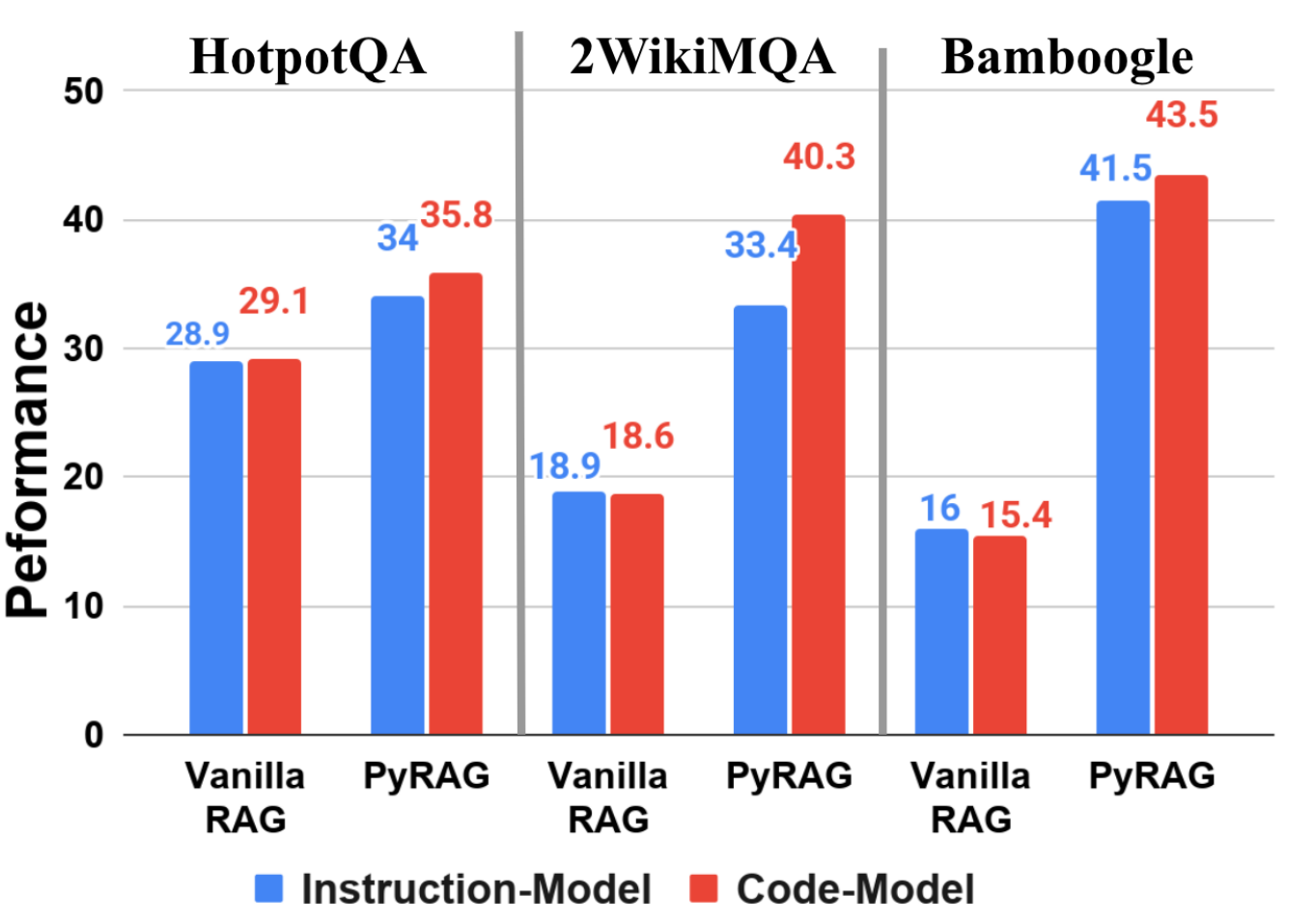}
    \caption{Code-specialized models show negligible advantage under Vanilla RAG but consistent gains under \pyrag, indicating that their benefit is task-dependent and emerges only when reasoning is formulated as program synthesis.}
    \label{fig:codemodel}
  \end{subfigure}

  \label{fig:ablation_and_codemodel}
\end{figure}

\paragraph{Progressive Component}

To understand the contribution of each component in \pyrag, we perform an ablation study that progressively introduces structure into the reasoning process. As shown in Figure \ref{fig:progressive_ablation}, we observe a consistent improvement from Vanilla RAG to \pyrag  across all three multi-hop benchmarks.

Introducing explicit decomposition (Decompose-only) yields modest gains over Vanilla RAG, indicating that breaking down complex questions into sub-queries already improves retrieval quality. However, representing the reasoning process as a structured plan (\pyrag w/o execution) leads to further improvements, suggesting that organizing intermediate steps—even without execution—helps guide the model toward more coherent reasoning.

The largest gains are achieved by \pyrag, which compiles and executes the generated plan as an executable program. This result highlights the importance of execution-based reasoning, where intermediate results are explicitly computed and passed across steps, rather than implicitly inferred.

\paragraph{Effect of Model Specialization.}
We further investigate whether PyRAG's gains arise from improved model capability
or from the proposed planning framework. As shown in Figure~\ref{fig:codemodel},
under Vanilla RAG, replacing the instruction-tuned model with a code-specialized
counterpart yields negligible differences across all three benchmarks (e.g.,
28.9 vs.\ 29.1 on HotpotQA, 18.9 vs.\ 18.6 on 2WikiMQA), indicating that
code-specialized models offer no general advantage in standard RAG. Under PyRAG,
however, the code-specialized model consistently outperforms the instruction-tuned
counterpart, with the gap widening on harder multi-hop benchmarks (+1.8 on HotpotQA,
+6.9 on 2WikiMQA, +2.0 on Bamboogle). Notably, even the instruction-model variant
of PyRAG already substantially outperforms Vanilla RAG, confirming that the gains
come primarily from structured planning, with code specialization providing
additional task-aligned leverage. This indicates that model capability and
reasoning interface must be co-designed: code models' strengths are realized
only when reasoning is explicitly formulated as program synthesis.

\subsection{Analysis}
\label{sec:analysis}

\paragraph{Efficiency Analysis}
We compare PyRAG against representative baselines in both EM and inference cost,
measured as the average number of LLM calls per query over 100 randomly sampled
HotpotQA queries; we select Search-R1 as the strongest RL-trained search agent
baseline.\footnote{For PyRAG, the Decompose and Plan stages are merged into a
  single LLM call; reported counts comprise this planning call plus all
  \texttt{answer()} invocations and any self-repair or adaptive-retrieval
re-executions.}

\begin{figure*}[t]
  \centering

  \begin{minipage}[c]{0.47\textwidth}
    \centering
    \includegraphics[width=\linewidth]{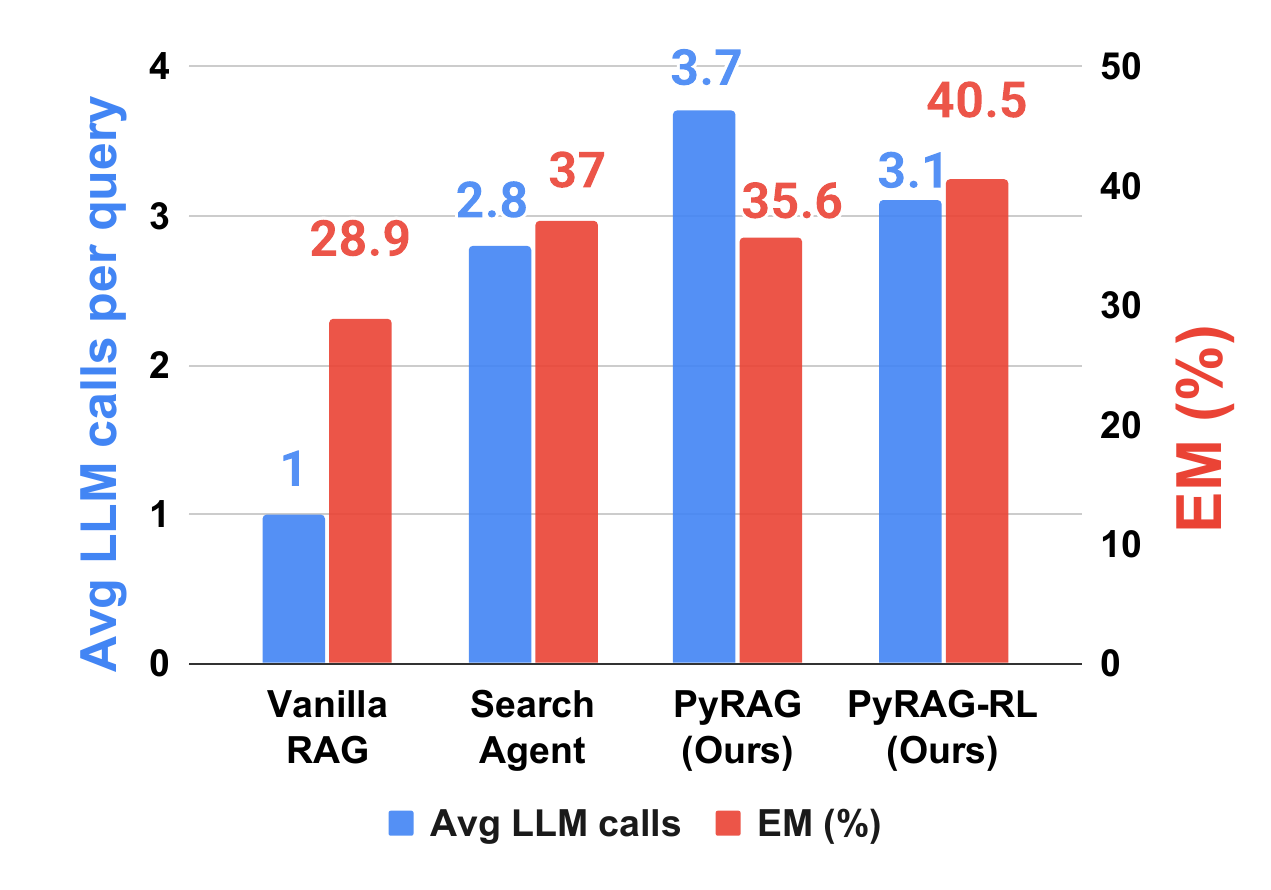}
    \subcaption{\pyrag achieves comparable EM to Search Agent with a modest increase in LLM calls, while \pyrag-RL attains the highest EM with fewer calls than the training-free \pyrag, indicating that RL fine-tuning produces more disciplined reasoning.}
    \label{fig:efficiency}
  \end{minipage}
  \hfill
  \begin{minipage}[c]{0.50\textwidth}
    \centering

    \includegraphics[width=\linewidth]{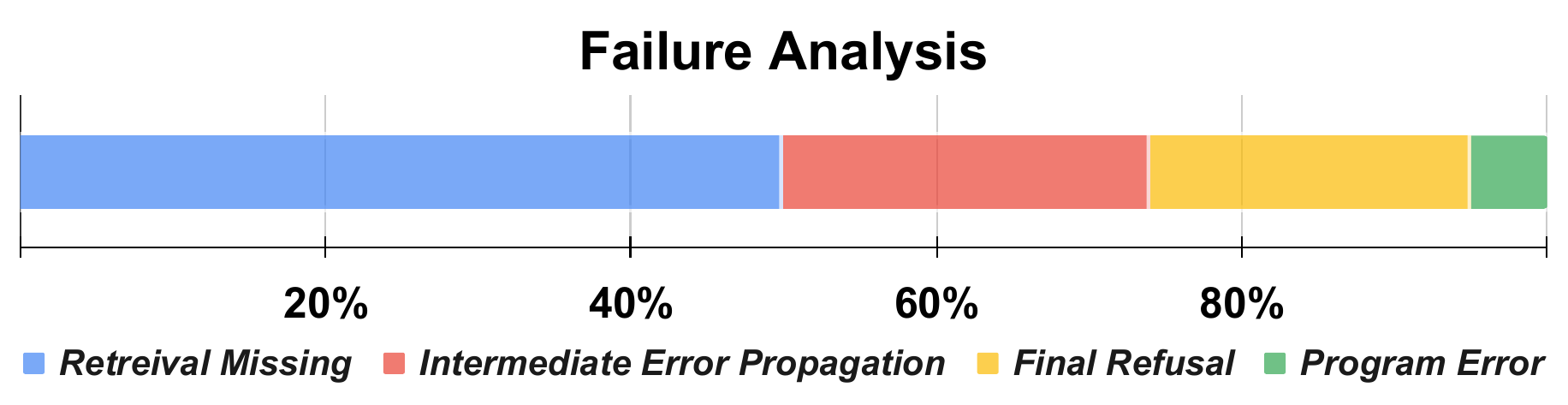}
    \subcaption{The answer agent accounts for $\sim$95\% of failures, while program errors contribute only $\sim$5\%, identifying the answer agent as the primary bottleneck.}
    \label{fig:failure_hotpot}

    \vspace{0.5em}

    \includegraphics[width=\linewidth]{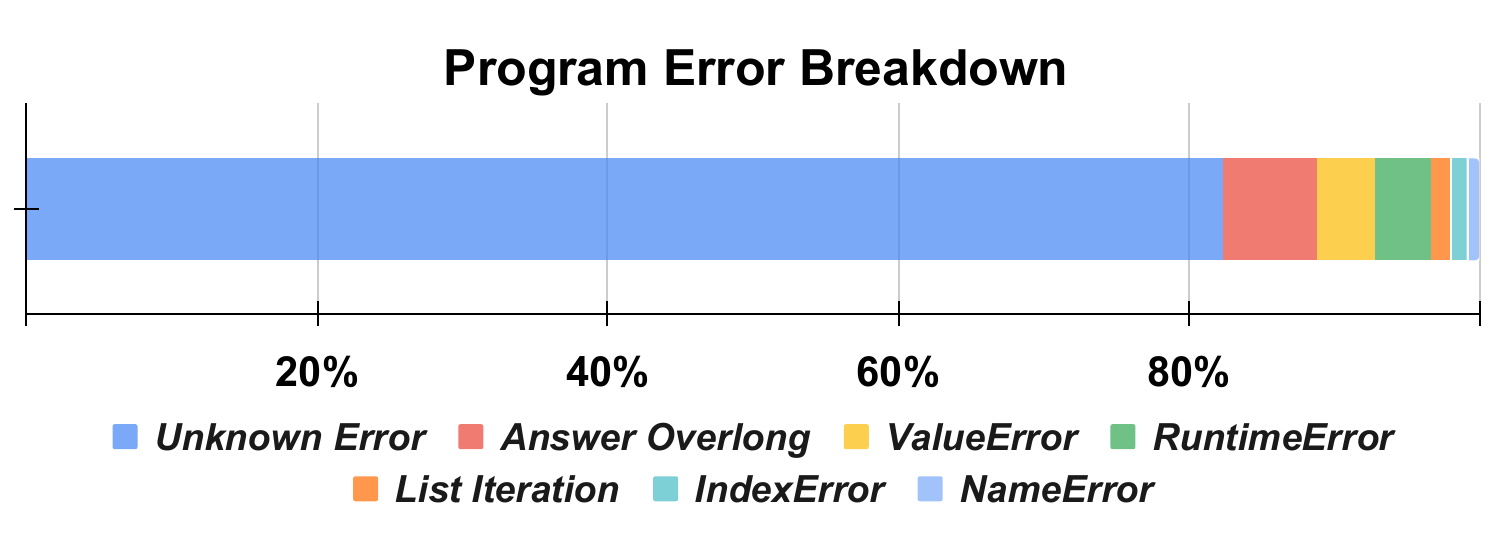}
    \caption{Among program errors, the dominant mode is Unknown Error, the program executes successfully but the answer agent fails to extract a grounded answer from retrieved evidence, rather than explicit runtime exceptions.}
    \label{fig:error_breakdown}
  \end{minipage}

  \label{fig:efficiency_error_analysis}
\end{figure*}

As shown in Figure~\ref{fig:efficiency}, Vanilla RAG is cheapest (one call) but
performs poorly on multi-hop questions, while Search-R1 improves accuracy through
unstructured iterative retrieval. PyRAG matches Search-R1's EM with a modest
3.7-call average, of which compiler-grounded self-repair triggers on $\sim$5\% of
queries and execution-driven adaptive retrieval on $\sim$20\%, indicating that
under-evidenced sub-steps rather than malformed programs are the primary driver
of re-executions. PyRAG-RL achieves the highest EM with even fewer calls (3.1 vs.\
3.7): RL fine-tuning produces more targeted queries and triggers both refinement
mechanisms less frequently as the policy becomes more reliable. Together, these
results indicate that the program-based structure assigns each LLM call a
well-defined role, yielding a better accuracy–cost trade-off than unstructured
iterative baselines.

\paragraph{Failure Analysis}
To understand the error sources of \pyrag, we manually categorize 100 randomly sampled incorrect predictions from HotpotQA. As shown in Figure~\ref{fig:failure_hotpot}, retrieval missing accounts for roughly half of all failures, identifying upstream retrieval recall as the dominant bottleneck. The next largest category is intermediate error propagation, where an uncertain sub-answer corrupts downstream steps (Failure F2), followed by final refusals where the answer agent declines despite the program executing as intended. Program errors contribute only $\sim$5\%, confirming that the planning agent reliably produces well-formed executable code.
We further characterize program errors among the same sampled cases (Figure~\ref{fig:error_breakdown}). The dominant mode is Unknown Error, in which the program executes without raising an exception but the answer agent returns a sentinel response (e.g., ``unknown'') because it fails to compose an answer from the retrieved evidence—a context-utilization issue rather than a Python-level fault. Genuine runtime exceptions (ValueError, TypeError, IndexError, NameError) together account for less than 20\% of program errors and are typically traceable to mismatched assumptions about retrieved string formats (e.g., Failure F5).

\paragraph{Case Study}
Due to space limitations, we defer the detailed case studies and qualitative examples to the appendix \ref{sec: case_study}.

%% file: related_work.tex
\section{Related Work}
\label{sec:related}
\paragraph{Multi-Hop Retrieval-Augmented Generation.}
Multi-hop QA requires chaining evidence across passages, which vanilla RAG~\cite{lewis2020retrieval} cannot handle in a single step. Prior work tackles this through iterative retrieve-and-reason prompting~\cite{wei2022chain, yao2023react, press2023measuring, trivedi2023interleaving, jiang2023active, shao2023enhancing, khattab2022demonstrate}, graph-based reasoning over retrieved content~\cite{edge2024local, gutierrez2024hipporag, chen2026pathrag, parekh2025structure}, and RL-trained search policies~\cite{jin2025searchr1, song2025r1searcher, zheng2025stepsearch, li2025search}. In all of these, the retrieval--reasoning interaction remains an implicit trajectory and error detection relies on LLM self-judgment. \pyrag instead represents the full pipeline as an executable program, making reasoning explicit and verifiable via compiler feedback.

\paragraph{Program-Guided Reasoning.}
Executable code has proven effective for reasoning over well-defined symbolic structures~\cite{gao2023pal, chen2022program, cheng2022binding, lyu2023faithful, pan2023logic, pan2023fact}, but these approaches assume the evidence is available \textit{a priori} in self-contained inputs. A complementary line, exemplified by DSPy~\cite{khattab2024dspy}, treats LM pipelines as compilable programs and optimizes their prompts. \pyrag targets a different setting—open-domain multi-hop QA where intermediate answers are unknown at synthesis time and later queries depend on earlier results, and contributes a concrete program-execution interface (see Appendix~\ref{app:related_work_extended} for extended discussion).

%% file: limitations.tex
\section{Limitations}
\label{sec:limitations}
While \pyrag demonstrates consistent gains across multi-hop benchmarks, our analysis (\ref{sec:analysis}) and case studies (Appendix~\ref{sec: case_study}) reveal several limitations that we believe are informative for future work.

\paragraph{Retrieval recall is the upstream bottleneck.}
Our failure analysis (Figure~\ref{fig:failure_hotpot}) shows that retrieval missing---where the retriever fails to surface gold evidence, accounts for roughly half of all incorrect predictions, making upstream retrieval recall the single largest source of failures. Although our adaptive retrieval mechanism mitigates this for sub-steps where the answer agent explicitly signals insufficient evidence, it cannot recover cases where retrieval silently returns plausible-looking but incorrect documents. Improving retrieval recall, for instance through query reformulation, learned retrievers, or hybrid sparse--dense retrieval, would yield the largest single accuracy gain and is largely orthogonal to \pyrag's contributions at the planning and execution layers.

\paragraph{Answer agents struggle to utilize retrieved context.}
Although our RL fine-tuning of the Answer Agent yields measurable improvements over the training-free variant, a substantial fraction of remaining failures still trace back to this stage, both as intermediate error propagation in Figure~\ref{fig:failure_hotpot} and as the dominant ``Unknown Error'' mode in Figure~\ref{fig:error_breakdown}, where the program executes successfully but the answer agent cannot ground a response in the retrieved passages. The executable interface successfully isolates failures to a specific stage, but the underlying difficulty, faithfully grounding answers in retrieved evidence and composing them across hops, is not eliminated by current RL objectives. Improving how language models exploit retrieved context, for instance through evidence-grounded training signals, calibrated uncertainty expression, or aggregation-aware objectives, is a key direction for future work.

\paragraph{Brittleness of sentinel-based adaptive retrieval.}
The execution-driven adaptive retrieval mechanism is triggered by string-level matching against sentinel responses such as ``unknown'' or ``cannot answer.'' Failure F2 illustrates a concrete weakness: when a sentinel value is interpolated into a downstream query as if it were content, retrieval errors silently propagate. A more robust design would replace string sentinels with structured return types or calibrated confidence signals.

\paragraph{Under-decomposition by the planner.}
Among program errors, the dominant mode is the silent single-retrieve case, where the planner emits syntactically valid code that issues only one \texttt{retrieve()} call for a question that requires multiple hops. This bypasses the reasoning chain entirely and is invisible to compiler-grounded self-repair, since no exception is raised. Plan complexity estimation, or an auxiliary objective penalizing under-decomposition, would be a natural complementary signal.

\section{Broader Impacts}
\label{sec:broader_impacts}
PyRAG aims to improve the factual grounding and interpretability of multi-hop question answering by replacing opaque natural-language reasoning trajectories with executable programs whose intermediate states are inspectable. We see two main positive impacts: (1) the inspectable execution trace lowers the barrier to auditing model behavior in knowledge-intensive applications, where silent reasoning errors or fabricated intermediate facts are otherwise difficult to localize; and (2) the program-execution interface decouples retrieval, computation, and aggregation, allowing deterministic operations (e.g., date arithmetic, boolean conjunction) to be handled outside the language model and reducing a known source of hallucination. At the same time, PyRAG inherits the risks of any retrieval-augmented system. Because final answers are grounded in retrieved passages, biases, factual errors, or under-representation of certain groups, languages, or domains in the underlying corpus can propagate into outputs while appearing well-supported by an inspectable trace—potentially lending unwarranted credibility to incorrect conclusions. The structured planning interface could also, in principle, be repurposed to automate the generation of seemingly evidence-backed but misleading content at scale. Finally, executing model-generated code introduces a standard but non-trivial security surface: deployments must sandbox the Python interpreter and restrict tool APIs to prevent untrusted programs from performing unintended actions. We restrict the runtime in our experiments to the two tool primitives retrieve and answer over a fixed Wikipedia corpus, and we encourage similar isolation in any downstream deployment.

%% file: tables/prompt.tex
This appendix lists the full prompts used by the three PyRAG agents
(Decompose, Plan, Answer) together with the two repair templates triggered by
compiler-grounded self-repair. Each prompt is reproduced verbatim from our
implementation; placeholders in braces (e.g.\ \textcolor{pyrag-placeholder}{\texttt{\{original\_query\}}})
are filled in at runtime. We use three colour codes throughout:
\textcolor{pyrag-prompt-sys}{\textbf{blue}} for system prompts that fix an
agent's role and output schema,
\textcolor{pyrag-prompt-user}{\textbf{green}} for user-side templates that
supply per-question context, and
\textcolor{pyrag-prompt-fix}{\textbf{orange}} for repair templates triggered
by execution failures.

\subsection{Decompose Agent}
\label{app:prompt-decompose}

The Decompose Agent maps the original multi-hop question $q$ to a list of
atomic, single-hop sub-queries $\mathbf{s}=[s_1,\dots,s_n]$. The system
prompt fixes a strict JSON-list output schema so the result can be parsed
deterministically and consumed by the Plan Agent without further
post-processing; the user prompt supplies the original question together
with a one-shot example that anchors the expected granularity (one
search-engine-answerable claim per item). Parsing failures trigger up to
three retries before falling back to using the original question as a
single-element list.

\begin{figure}[h]
  \centering
  \begin{tcolorbox}[
      colback=pyrag-prompt-sys-bg, colframe=pyrag-prompt-sys,
      title={\textbf{Decompose Agent}\hfill\textsc{System Prompt}},
      fonttitle=\normalsize, boxrule=0.6pt, arc=2pt,
      left=4pt, right=4pt,
    ]
\begin{lstlisting}[style=pyragprompt]
You are a question decomposition agent for multi-hop QA.
Break a complex question into a minimal list of atomic single-hop
sub-queries. Each sub-query should be independently answerable by a
search engine. Return a JSON list of strings ONLY. No explanation,
no markdown, no extra text.
\end{lstlisting}
  \end{tcolorbox}
  \vspace{-6pt}
  \begin{tcolorbox}[
      colback=pyrag-prompt-user-bg, colframe=pyrag-prompt-user,
      title={\textbf{Decompose Agent}\hfill\textsc{User Template}},
      fonttitle=\normalsize, boxrule=0.6pt, arc=2pt,
      left=4pt, right=4pt,
    ]
\begin{lstlisting}[style=pyragprompt]
Original question:
{query}

Return a JSON list of atomic sub-queries. Example:
["Who directed Inception?",
 "Who directed Jurassic Park?",
 "When was Christopher Nolan born?"]
\end{lstlisting}
  \end{tcolorbox}
  \caption{Prompts used by the Decompose Agent. The system prompt enforces a
  parseable JSON-list contract; the user template supplies the question and a
  one-shot example fixing sub-query granularity.}
  \label{fig:prompt-decompose}
\end{figure}

\subsection{Plan Agent}
\label{app:prompt-plan}

The Plan Agent is the core of PyRAG: given the original question $q$ and the
decomposed sub-queries $\mathbf{s}$, it synthesises an executable Python
program over the two tool primitives \texttt{retrieve(query)} and
\texttt{answer(query, docs)}. The system prompt
(Figure~\ref{fig:prompt-plan-sys}) codifies the executable interface as a
contract: it specifies the exact function signatures, including a
no-\texttt{docs} aggregation mode for the final synthesis call; the data-flow
discipline that intermediate results must be bound to identifiers and reused
via f-strings rather than re-derived; and the two-part synthesis format
``\emph{Given: \textless{}facts\textgreater. Answer the question:
\textless{}original question verbatim\textgreater}'' that prevents an
intermediate answer from leaking into the question template. The user
prompt instantiates this contract for the specific query and supplies a
worked one-shot example demonstrating variable threading across hops.

\begin{figure}[h]
  \centering
  \begin{tcolorbox}[
      colback=pyrag-prompt-sys-bg, colframe=pyrag-prompt-sys,
      title={\textbf{Plan Agent}\hfill\textsc{System Prompt}},
      fonttitle=\normalsize, boxrule=0.6pt, arc=2pt,
      left=4pt, right=4pt,
    ]
\begin{lstlisting}[style=pyragprompt]
You are a planning agent that writes Python code to answer questions
via RAG.

You have exactly two functions available:

  retrieve(query: str) -> List[str]
      Searches a retrieval system and returns a list of relevant
      document strings. Do NOT pass a topk argument; the retrieval
      count is controlled externally.

  answer(query: str, docs: List[str]) -> str
      Calls an LLM to answer a question given a list of documents.
      Returns ONLY the short text inside the model's <answer>...
      </answer> block (the runtime extracts it). Use that return
      value directly in f-strings -- it is a short phrase, not the
      full model reply.

  answer(query: str) -> str  (no docs -- aggregation / synthesis mode)
      Same as above but with NO documents. The model answers using
      ONLY the information already present in the query string. Use
      this for the FINAL aggregation step.

Rules:
- Use retrieve() + answer() step by step to answer each sub-query.
- If you call retrieve() for a sub-question, pass that return value
  into the matching answer() for that sub-question.
- Build on intermediate answers by interpolating them into later
  queries using f-strings.
- Variable names MUST be valid Python identifiers: only letters,
  digits, and underscores, NO spaces.
- Every variable you reference MUST be assigned BEFORE it is used.
- Do NOT parse answer() return values in generated code (no .split(),
  regex, or indexing). The runtime already returns the short <answer>
  span.
- If you use a for/if/while block that might conditionally set
  final_answer, initialise final_answer = "" BEFORE the block.
- The last line of your code MUST be:
      final_answer = answer(<synthesis question>)
  Do NOT pass docs to this final answer() call -- aggregation only.
- NEVER use an f-string or string concatenation as final_answer.
  final_answer MUST ALWAYS be the return value of an answer() call.
- CRITICAL: The <synthesis question> must use this TWO-PART format:
    "Given: <fact1>, <fact2>, ...
     Answer the question: <ORIGINAL QUESTION>"
  where the first part lists intermediate results as background facts
  and the second part REPEATS the ORIGINAL question VERBATIM after
  "Answer the question:". Do NOT rephrase the original question or
  embed intermediate answers INTO the question itself.
  BAD:  f"Which American director, {name}, who is {nationality},
         hosted ...?"  (answer leaks into the question)
  GOOD: f"Given: {name} hosted the event, {name} is {nationality}.
         Answer the question: Which American director hosted ...?"
- Valid Python 3 only: use ASCII double or single quotes for strings;
  never put backticks around variable names or expressions.
- Return ONLY the Python code, no explanation, no imports, no markdown.
\end{lstlisting}
  \end{tcolorbox}
  \caption{Plan Agent system prompt. Codifies the executable interface as a
  contract --- function signatures, data-flow discipline, and the two-part
  synthesis format that prevents answer leakage into the final question.}
  \label{fig:prompt-plan-sys}
\end{figure}

\begin{figure}[h]
  \centering
  \begin{tcolorbox}[
      colback=pyrag-prompt-user-bg, colframe=pyrag-prompt-user,
      title={\textbf{Plan Agent}\hfill\textsc{User Template}},
      fonttitle=\normalsize, boxrule=0.6pt, arc=2pt,
      left=4pt, right=4pt,
    ]
\begin{lstlisting}[style=pyragprompt]
Original question:
{original_query}

Sub-queries to resolve:
{sub_queries}

Reference example (do NOT copy, write code for the actual question
above):
{CODE_EXAMPLE}

Now write the Python code for the original question.
End with: final_answer = answer(f"Given: <facts>.
                                  Answer the question: {original_query}")
\end{lstlisting}
  \end{tcolorbox}
  \vspace{-6pt}
  \begin{tcolorbox}[
      colback=pyrag-prompt-user-bg, colframe=pyrag-prompt-user,
      title={\textbf{Plan Agent}\hfill\textsc{One-Shot Example (\texttt{CODE\_EXAMPLE})}},
      fonttitle=\normalsize, boxrule=0.6pt, arc=2pt,
      left=4pt, right=4pt,
    ]
\begin{lstlisting}[style=pyragcode]
# Example for: "Who was born earlier, the director of
# Inception or Jurassic Park?"
docs1 = retrieve("Who directed Inception?")
director1 = answer("Who directed Inception?", docs1)

docs2 = retrieve("Who directed Jurassic Park?")
director2 = answer("Who directed Jurassic Park?", docs2)

docs3 = retrieve(f"When was {director1} born?")
birth1 = answer(f"When was {director1} born?", docs3)

docs4 = retrieve(f"When was {director2} born?")
birth2 = answer(f"When was {director2} born?", docs4)

# ALWAYS end with answer() -- list facts first,
# then repeat the ORIGINAL question verbatim
final_answer = answer(
    f"Given: {director1} directed Inception and was born {birth1}, "
    f"{director2} directed Jurassic Park and was born {birth2}. "
    f"Answer the question: Who was born earlier, the director of "
    f"Inception or Jurassic Park?"
)
\end{lstlisting}
  \end{tcolorbox}
  \caption{Plan Agent user-side context. Top: the per-question template
  filled with the original query, the decomposed sub-queries, and the
  one-shot \texttt{CODE\_EXAMPLE}. Bottom: the example itself, demonstrating
  variable threading across hops and the two-part synthesis call.}
  \label{fig:prompt-plan-user}
\end{figure}

\subsubsection{Self-Repair Templates}
\label{app:prompt-plan-repair}

PyRAG's compiler-grounded self-repair operates at two granularities.
\emph{Syntax-level} feedback (Figure~\ref{fig:prompt-plan-syntax}) is
triggered when the model's output fails to compile under
\texttt{compile(\dots,\,\textquotesingle exec\textquotesingle)}; the failed
snippet and the parser's error location are returned to the same generation
call, up to three retries. \emph{Runtime-level} feedback
(Figure~\ref{fig:prompt-plan-fix}) is triggered after a successful compile
when the program raises a Python exception during execution; the original
question, the failing program, and the traceback are surfaced back to the
Plan Agent, which produces a corrected program for the runtime to
re-execute. Both templates re-iterate the contract violations most commonly
responsible for failure (uninitialised \texttt{final\_answer}, parsed
\texttt{answer()} return values, missing \texttt{docs} arguments) so that
repair is grounded in deterministic compiler signals rather than the
model's self-judgement.

\begin{figure}[h]
  \centering
  \begin{tcolorbox}[
      colback=pyrag-prompt-fix-bg, colframe=pyrag-prompt-fix,
      title={\textbf{Plan Agent}\hfill\textsc{Runtime-Error Repair Template}},
      fonttitle=\normalsize, boxrule=0.6pt, arc=2pt,
      left=4pt, right=4pt,
    ]
\begin{lstlisting}[style=pyragprompt]
The following Python code was generated to answer the question but
raised a runtime error. Fix the code so it runs correctly. Return
ONLY the corrected Python code.

=== Original question ===
{original_query}

=== Failing code ===
{failed_code}

=== Runtime error ===
{error_msg}

=== Fix instructions ===
- Do not parse answer() return values in generated code; the runtime
  returns short <answer> text.
- Every variable you reference must be explicitly assigned earlier.
- If retrieve() was called, pass its docs into answer(); do not use
  answer(..., []) for that step.
- Initialise final_answer = "" before any loop/conditional that
  might set it.
- The last line MUST be: final_answer = answer(<synthesis question>)
  -- call answer() with NO docs to aggregate all intermediate
  results. NEVER use an f-string or string as final_answer directly.
\end{lstlisting}
  \end{tcolorbox}
  \caption{Runtime-level self-repair template. Triggered when an executed
  program raises a Python exception; the traceback and the original
  question are surfaced back as deterministic, grounded feedback signals.}
  \label{fig:prompt-plan-fix}
\end{figure}

\begin{figure}[h]
  \centering
  \begin{tcolorbox}[
      colback=pyrag-prompt-fix-bg, colframe=pyrag-prompt-fix,
      title={\textbf{Plan Agent}\hfill\textsc{Syntax-Error Repair Template}},
      fonttitle=\normalsize, boxrule=0.6pt, arc=2pt,
      left=4pt, right=4pt,
    ]
\begin{lstlisting}[style=pyragprompt]
---
Your previous output is NOT valid Python (syntax error).
Details: {error_detail}

Fix it. Use only ASCII ' or " for strings; never wrap variable names
in backticks; output code only.

Previous attempt:
```python
{failed_code}
```
Return ONLY corrected Python code.
\end{lstlisting}
  \end{tcolorbox}
  \caption{Syntax-level self-repair template. Triggered when the generated
  code fails to compile; the parser's error location is fed back to the
  same generation call.}
  \label{fig:prompt-plan-syntax}
\end{figure}

\subsection{Answer Agent}
\label{app:prompt-answer}

The Answer Agent is invoked once per \texttt{answer(query, docs)} call in the
executed program and operates in two distinct modes that share an identical
\texttt{<redacted\_thinking>}\,/\,\texttt{<answer>} output schema. In
\emph{evidence mode} (Figure~\ref{fig:prompt-answer-docs}), the agent
receives a sub-query together with retrieved passages, must answer using
only those passages, and must cite each used passage inline as
``\texttt{Doc [i]}''. The schema fixes type-matching between the question
and the answer span (a \emph{who} question must return a name, not a date)
and reserves the literal token ``\texttt{unknown}'' as a sentinel for
under-evidenced steps, which directly drives the execution-driven adaptive
retrieval mechanism described in Section~2.6 of the main text. In
\emph{aggregation mode} (Figure~\ref{fig:prompt-answer-nodocs}), no
documents are supplied; the prompt instead relies on the two-part
\emph{Given:\,\dots\,Answer the question:\,\dots} template emitted by the
Plan Agent, and the agent composes the final answer from the supplied
facts. The aggregation prompt explicitly forbids yes/no responses to
\emph{wh}-questions, addressing the failure mode in which the model
otherwise treats the synthesis call as fact verification. Retrieved
documents are formatted as
\verb|[Doc 1]\n<text>\n\n[Doc 2]\n<text>\n\n...| so that inline citation
indices are unambiguous.

\begin{figure}[h]
  \centering
  \begin{tcolorbox}[
      colback=pyrag-prompt-sys-bg, colframe=pyrag-prompt-sys,
      title={\textbf{Answer Agent}\hfill\textsc{Evidence Mode (with documents)}},
      fonttitle=\normalsize, boxrule=0.6pt, arc=2pt,
      left=4pt, right=4pt,
    ]
\begin{lstlisting}[style=pyragprompt]
You are given a question and retrieved documents.
You MUST answer using ONLY information from the retrieved documents.
Even for yes/no questions, decide yes or no by reasoning from facts
in the documents.

Output format (STRICT):
<thinking> ... </thinking>
<answer> ... </answer>

Evidence citation rule:
- Whenever you use evidence from the documents in your reasoning,
  you MUST cite it inline as Doc [i] (matching the document indices
  shown in the retrieved block, e.g. [Doc 1] -> Doc [1]).
- Only cite documents that are actually relevant.
- Keep <thinking> concise (1-3 sentences).

Answer rules:
- The <answer> should be a short phrase, preferably taken directly
  from the documents when possible.
- Match the answer TYPE to the QUESTION:
    WHO / which person / born first / earlier
        -> a person's NAME in <answer>, not only a date;
    WHEN -> date or time;
    yes/no -> exactly yes / no / unknown when the documents do not
              support a definite answer.
- Do NOT output anything outside <thinking> and <answer>.

Example (do NOT copy the content, only follow the style):
<thinking>Doc [1] states that Future Ted serves as the
narrator, and Doc [4] confirms the voice actor.</thinking>
<answer> Ted Mosby </answer>
\end{lstlisting}
  \end{tcolorbox}
  \caption{Answer Agent system prompt --- \emph{evidence mode}. Used when at
  least one retrieved passage is supplied. The schema fixes question-type
  matching and reserves ``\texttt{unknown}'' as the sentinel that drives
  adaptive retrieval.}
  \label{fig:prompt-answer-docs}
\end{figure}

\begin{figure}[h]
  \centering
  \begin{tcolorbox}[
      colback=pyrag-prompt-sys-bg, colframe=pyrag-prompt-sys,
      title={\textbf{Answer Agent}\hfill\textsc{Aggregation Mode (no documents)}},
      fonttitle=\normalsize, boxrule=0.6pt, arc=2pt,
      left=4pt, right=4pt,
    ]
\begin{lstlisting}[style=pyragprompt]
There are NO retrieved documents. The question text itself contains
background facts (after 'Given:') and the actual question to answer
(after 'Answer the question:').
You MUST use the provided facts to answer the ACTUAL QUESTION.

CRITICAL: Your job is to ANSWER the question, NOT to confirm whether
the facts are true.
- If the question asks WHO / WHICH person -> reply with a person's NAME.
- If the question asks WHEN -> reply with a date or time.
- If the question asks WHERE -> reply with a place.
- ONLY answer yes/no when the question is explicitly a yes/no question
  (e.g. 'Are both ...?', 'Is it true ...?').
- NEVER answer 'yes' or 'no' to a WHO/WHICH/WHEN/WHERE question.

Output format (STRICT):
<thinking> ... </thinking>
<answer> ... </answer>

- In <thinking>, identify the actual question type and
  combine the given facts to produce the answer (1-2 sentences).
- The <answer> must directly answer the question -- a name, date,
  place, etc. -- NOT 'yes' or 'no' unless the question is truly
  yes/no.
- Do NOT output anything outside <thinking> and <answer>.
\end{lstlisting}
  \end{tcolorbox}
  \caption{Answer Agent system prompt --- \emph{aggregation mode}. Used when
  the \texttt{docs} argument is empty, i.e.\ in the final synthesis call.
  Forbids yes/no responses to \emph{wh}-questions, eliminating the failure
  mode where the synthesis call collapses into fact verification.}
  \label{fig:prompt-answer-nodocs}
\end{figure}

%% file: tables/case_study.tex
\begin{figure}[t]
  \centering
  \begin{tcolorbox}[
      colback=pyrag-bg-good, colframe=pyrag-good,
      title={\textbf{Case~A}\hfill
      {\textcolor{pyrag-good}{\ding{51}}~Correct}},
      fonttitle=\normalsize, boxrule=0.6pt, arc=2pt, left=4pt, right=4pt,
    ]
    \textbf{Question:} \emph{2014 S/S is the debut album of a South Korean boy group that was formed by who?}\\[2pt]
    \textbf{Gold:} \texttt{YG Entertainment} \hfill \textbf{Predicted:} \texttt{YG Entertainment}
    \tcblower
\begin{lstlisting}[style=pyragcode]
# Step 1-2: identify the boy group
ctx_grp = retrieve("2014 S/S debut South Korean boy group", topk=5)
group   = answer("Which group is the album by?", ctx_grp)
# group = "WINNER"

# Step 3-4: the next query *uses* the bound variable `group`
ctx_co  = retrieve(f"Which company formed {group}?", topk=5)
company = answer(f"Which company formed {group}?", ctx_co)
# company = "YG Entertainment"

# Step 5: final aggregation is itself an answer(...) call whose query
#         is built from the two intermediate variables
final = answer(
    f"Given: {group} is a South Korean boy group; "
    f"{company} formed {group}. "
    f"Answer the question: 2014 S/S is the debut album "
    f"of a South Korean boy group that was formed by who?",
    ctx_co,
)
# final -> "YG Entertainment"   [ok]
\end{lstlisting}
    
  \end{tcolorbox}
  \caption{A representative correct example. Variables produced at one step are explicitly consumed by subsequent calls through string interpolation.}
  \label{fig:case-bridge}
\end{figure}

\begin{figure}[t]
  \centering
  \begin{tcolorbox}[
      colback=pyrag-bg-good, colframe=pyrag-good,
      title={\textbf{Case~B}\hfill
      {\textcolor{pyrag-good}{\ding{51}}~Correct}},
      fonttitle=\normalsize, boxrule=0.6pt, arc=2pt, left=4pt, right=4pt,
    ]
    \textbf{Question:} \emph{Which other Mexican Formula One race car driver has held the podium besides the Force India driver born in 1990?}\\[2pt]
    \textbf{Gold:} \texttt{Pedro Rodríguez} \hfill \textbf{Predicted:} \texttt{Pedro Rodríguez}
    \tcblower
\begin{lstlisting}[style=pyragcode]
# Step 1-2: a known Mexican F1 podium driver
ctx_a   = retrieve("Mexican F1 driver who held the podium", topk=5)
known   = answer("Who is the Mexican F1 podium driver?", ctx_a)
# known = "Sergio P\'erez"

# Step 3-4: try to identify the Force India 1990-born driver
ctx_b   = retrieve("Force India driver born in 1990", topk=5)
exclude = answer("Which Force India driver was born in 1990?", ctx_b)
# exclude = "unknown"      [warn] insufficient evidence
\end{lstlisting}
\begin{lstlisting}[style=pyragcode_ok]
# Step 3-4: executor reissues a broader retrieval (adaptive re-retrieval)
ctx_b   = retrieve("Force India driver born in 1990", topk=10)
exclude = answer("Which Force India driver was born in 1990?", ctx_b)
# exclude = "Pedro Rodr\'iguez and Sergio P\'erez"
\end{lstlisting}
\begin{lstlisting}[style=pyragcode]
# Step 5: the final aggregation is again an answer(...) call,
#         whose query templates in the previously bound variables
final = answer(
    f"Given: {known} is a Mexican F1 podium driver; "
    f"the Mexican F1 podium drivers are {exclude}. "
    f"Which other Mexican F1 podium driver is there besides {known}?",
    ctx_c,
)
# final -> "Pedro Rodr\'iguez"   [ok]
\end{lstlisting}
    
  \end{tcolorbox}
  \caption{When Step~4 returns the sentinel \texttt{"unknown"}, execution-guided refinement triggers a broader re-retrieval (Step~5--6, highlighted). The plan structure is preserved; only the under-evidenced sub-step is repaired. Adaptive retrieval recovers from an under-evidenced sub-step without modifying the overall plan, illustrating the benefit of execution-grounded refinement.}
  \label{fig:case-selfcorrect}
\end{figure}

\begin{figure}[t]
  \centering
  \begin{tcolorbox}[
      colback=pyrag-bg-good, colframe=pyrag-good,
      title={\textbf{Case~C}\hfill
      {\textcolor{pyrag-good}{\ding{51}}~Correct}},
      fonttitle=\normalsize, boxrule=0.6pt, arc=2pt, left=4pt, right=4pt,
    ]
    \textbf{Question:} \emph{Are Freakonomics and In the Realm of the Hackers both American documentaries?}\\[2pt]
    \textbf{Gold:} \texttt{no} \hfill \textbf{Predicted:} \texttt{No}
    \tcblower
\begin{lstlisting}[style=pyragcode]
films      = ["Freakonomics", "In the Realm of the Hackers"]
predicates = ["is a documentary", "is American"]

# Two predicates per film -> grid of 4 yes/no probes
flags = {f: {} for f in films}
for film in films:
    for pred in predicates:
        ctx               = retrieve(f"Is {film} {pred}?", topk=5)
        flags[film][pred] = answer(f"Is {film} {pred}?", ctx)

# flags = {
#   "Freakonomics":              {"is a documentary": "no",  "is American": "yes"},
#   "In the Realm of the Hackers":{"is a documentary": "yes", "is American": "no"},
# }

# Boolean structure handled in Python -- not by the answer agent
def qualifies(d):
    return all(v.lower() == "yes" for v in d.values())

if all(qualifies(flags[f]) for f in films):
    final = "yes"
else:
    final = "no"
# final -> "no"   (Freakonomics fails the documentary check;
#                 In the Realm of the Hackers fails the American check)   [ok]
\end{lstlisting}
  \end{tcolorbox}
  \caption{Boolean conjunction over a 2$\times$2 grid of predicates. The plan reduces a ``both X and Y'' question to a Cartesian grid of yes/no probes whose conjunction is decided by the Python keyword \texttt{all}. The boolean structure is enforced by the program; the answer agent never has to perform multi-clause logical reasoning over a free-form prompt. The decision rule is expressed as a Python expression rather than delegated to the answer agent.}
  \label{fig:case-bool}
\end{figure}

\begin{figure}[t]
  \centering
  \begin{tcolorbox}[
      colback=pyrag-bg-good, colframe=pyrag-good,
      title={\textbf{Case~D}\hfill
      {\textcolor{pyrag-good}{\ding{51}}~Correct}},
      fonttitle=\normalsize, boxrule=0.6pt, arc=2pt, left=4pt, right=4pt,
    ]
    \textbf{Question:} \emph{How old was Virginia Bruce when she starred in Let Freedom Ring?}\\[2pt]
    \textbf{Gold:} \texttt{29} \hfill \textbf{Predicted:} \texttt{29}
    \tcblower
\begin{lstlisting}[style=pyragcode]
import re

# Two retrievals supply the numeric inputs
ctx_film = retrieve("In what year did Virginia Bruce star in Let Freedom Ring?", topk=5)
year_str = answer("In what year did Virginia Bruce star in Let Freedom Ring?", ctx_film)
# year_str = "1939"

ctx_born = retrieve("In what year was Virginia Bruce born?", topk=5)
born_str = answer("In what year was Virginia Bruce born?", ctx_born)
# born_str = "1910"

# Cast text -> int and compute the answer in Python; no further LLM call needed
year_film = int(re.search(r"\d{4}", year_str).group())
year_born = int(re.search(r"\d{4}", born_str).group())
final     = year_film - year_born
# final -> 29   [ok]
\end{lstlisting}
  \end{tcolorbox}
  \caption{Arithmetic over retrieved values. The final answer is not contained in any retrieved document and cannot be produced by retrieve+answer alone---it is the \emph{difference} of two retrieved years. PyRAG handles this by lifting the retrieved strings into Python integers via \texttt{int(...)} and computing the subtraction deterministically. An LLM doing arithmetic on natural-language dates inside a free-form prompt is a known source of error; here the computation is moved outside the model entirely. PyRAG separates retrieval (handled by tools) from computation (handled by Python), giving deterministic numeric answers without relying on LLM mental arithmetic.}
  \label{fig:case-arith}
\end{figure}

\begin{figure}[t]
  \centering
  \begin{tcolorbox}[
      colback=pyrag-bg-bad, colframe=pyrag-bad,
      title={\textbf{Failure~F1}\hfill
      {\textcolor{pyrag-bad}{\ding{55}}~Incorrect}},
      fonttitle=\normalsize, boxrule=0.6pt, arc=2pt, left=4pt, right=4pt,
    ]
    \textbf{Question:} \emph{Aside from the Apple Remote, what other device can control the program Apple Remote was originally designed to interact with?}\\[2pt]
    \textbf{Gold:} \texttt{keyboard function keys} \hfill \textbf{Predicted:} \texttt{iPod Touch, iPad}~\textcolor{pyrag-bad}{\ding{55}}
    \tcblower
\begin{lstlisting}[style=pyragcode]
# Step 1-2: correctly identify the program
ctx_p   = retrieve("Program the Apple Remote was designed for", topk=5)
program = answer("What program is Apple Remote designed for?", ctx_p)
# program = "Front Row"   [ok]
\end{lstlisting}
\begin{lstlisting}[style=pyragcode_warn]
# Step 3-4: planner SHOULD use `program` here, but silently rebinds to "iTunes"
ctx_d  = retrieve("What other devices can control iTunes?", topk=5)   # [warn] entity drift
device = answer("What other devices can control iTunes?", ctx_d)
# device = "iPod Touch, iPad"
\end{lstlisting}
\begin{lstlisting}[style=pyragcode]
# Step 5: final answer(...) propagates the corrupted variable
final = answer(
    f"Given: Apple Remote was designed for {program}; "
    f"other devices that can control {program} include {device}. "
    f"Aside from the Apple Remote, what other device can control {program}?",
    ctx_d,
)
# final -> "iPod Touch, iPad"   [wrong]
\end{lstlisting}
  \end{tcolorbox}
  \caption{Decomposition-stage entity drift. The Step~3 query should have been \texttt{f"...control \{program\}?"}, but the planning agent emitted the literal string \texttt{"iTunes"} instead. Because the plan exposes variables as first-class objects, the drift is precisely localizable to the planning stage---a free-form CoT trace would mix this error into surrounding reasoning text. The executable trace makes the failure point unambiguous.}
  \label{fig:fail-decomp}
\end{figure}

\begin{figure}[t]
  \centering
  \begin{tcolorbox}[
      colback=pyrag-bg-bad, colframe=pyrag-bad,
      title={\textbf{Failure~F2}\hfill
      {\small\textsc{HotpotQA}\,id=47 \quad \textcolor{pyrag-bad}{\ding{55}}~Incorrect}},
      fonttitle=\normalsize, boxrule=0.6pt, arc=2pt, left=4pt, right=4pt,
    ]
    \textbf{Question:} \emph{According to the 2001 census, what was the population of the city in which Kirton End is located?}\\[2pt]
    \textbf{Gold:} \texttt{35,124} \hfill \textbf{Predicted:} \texttt{12,790}~\textcolor{pyrag-bad}{\ding{55}}
    \tcblower
\begin{lstlisting}[style=pyragcode]
# Step 1-2: locality lookup succeeds
ctx_loc  = retrieve("Where is Kirton End located?", topk=5)
locality = answer("Where is Kirton End located?", ctx_loc)
# locality = "Kirton, Lincolnshire"

# Step 3-4: the next hop (city containing the locality) FAILS
ctx_city = retrieve(f"In what city is {locality}?", topk=10)
city     = answer(f"In what city is {locality}?", ctx_city)
# city = "unknown"   [warn]
\end{lstlisting}
\begin{lstlisting}[style=pyragcode_warn]
# Step 5-6: sentinel value "unknown" is interpolated as a *content* string
ctx_pop = retrieve(f"Population of {city} in the 2001 census", topk=5)   # [warn]
pop     = answer(f"Population of {city} in the 2001 census?", ctx_pop)
# pop = "12,790"   (population of an unrelated locality)
\end{lstlisting}
\begin{lstlisting}[style=pyragcode]
# Step 7: final answer(...) silently consumes the corrupted hop
final = answer(
    f"Given: Kirton End is located in {locality}; "
    f"the containing city is {city}; "
    f"the 2001 census population of {city} was {pop}. "
    f"What was the 2001 census population of the city in which Kirton End is located?",
    ctx_pop,
)
# final -> "12,790"   [wrong]
\end{lstlisting}
  \end{tcolorbox}
  \caption{Retrieval failure that propagates because a sentinel value is treated as a content string. The string \texttt{"unknown"} returned at Step~4 is a sentinel meaning ``no evidence,'' but the plan treats it as a normal value and interpolates it into Step~5's query. This points to a concrete fix: hops whose answer matches the \texttt{"unknown"} sentinel should branch into guarded fallback rather than continuing the data-flow chain. The executable trace localizes this to a single edge in the data dependency graph.}
  \label{fig:fail-retrieval}
\end{figure}

\begin{figure}[t]
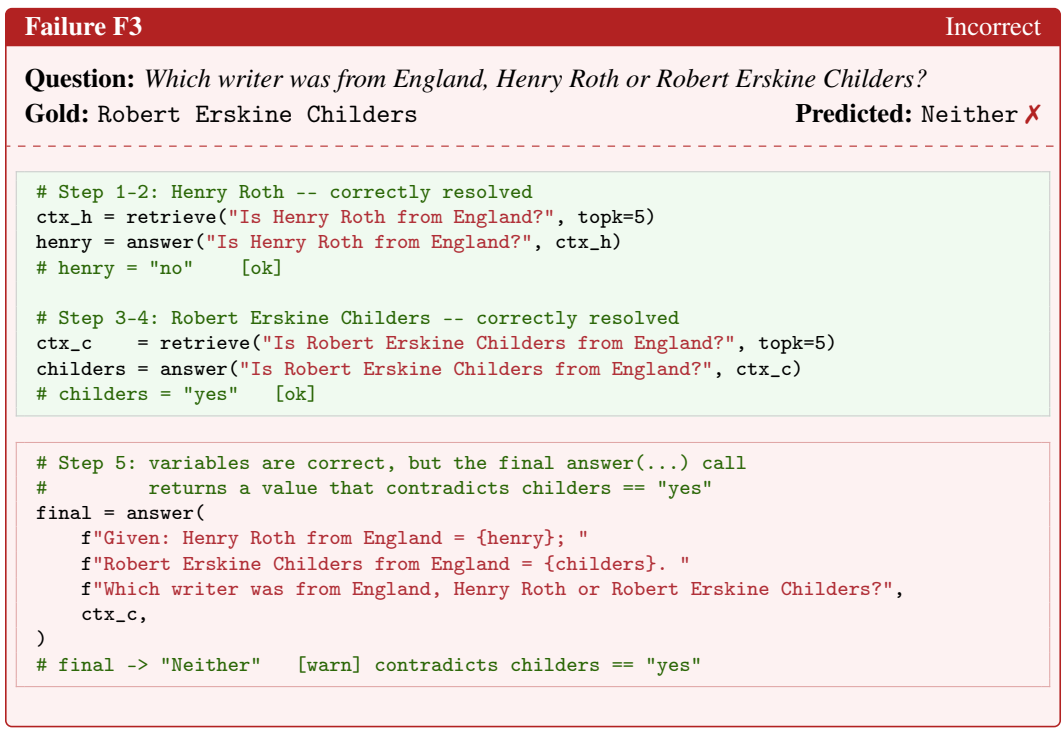

  \centering
  \begin{tcolorbox}[
      colback=pyrag-bg-bad, colframe=pyrag-bad,
      title={\textbf{Failure~F3}\hfill
      {\textcolor{pyrag-bad}{\ding{55}}~Incorrect}},
      fonttitle=\normalsize, boxrule=0.6pt, arc=2pt, left=4pt, right=4pt,
    ]
    \textbf{Question:} \emph{Which writer was from England, Henry Roth or Robert Erskine Childers?}\\[2pt]
    \textbf{Gold:} \texttt{Robert Erskine Childers} \hfill \textbf{Predicted:} \texttt{Neither}~\textcolor{pyrag-bad}{\ding{55}}
    \tcblower
\begin{lstlisting}[style=pyragcode]
# Step 1-2: Henry Roth -- correctly resolved
ctx_h = retrieve("Is Henry Roth from England?", topk=5)
henry = answer("Is Henry Roth from England?", ctx_h)
# henry = "no"    [ok]

# Step 3-4: Robert Erskine Childers -- correctly resolved
ctx_c    = retrieve("Is Robert Erskine Childers from England?", topk=5)
childers = answer("Is Robert Erskine Childers from England?", ctx_c)
# childers = "yes"   [ok]
\end{lstlisting}
\begin{lstlisting}[style=pyragcode_warn]
# Step 5: variables are correct, but the final answer(...) call
#         returns a value that contradicts childers == "yes"
final = answer(
    f"Given: Henry Roth from England = {henry}; "
    f"Robert Erskine Childers from England = {childers}. "
    f"Which writer was from England, Henry Roth or Robert Erskine Childers?",
    ctx_c,
)
# final -> "Neither"   [warn] contradicts childers == "yes"
\end{lstlisting}
  \end{tcolorbox}
  \caption{Final aggregation misreads its own variable bindings. Written as a Python program, the failure is sharply localized: every retrieved variable holds the right value, yet the final \texttt{answer(...)} call returns \texttt{"Neither"}. The bug is therefore neither retrieval nor variable binding, it is the answer agent misreading its own variable bindings inside the final aggregation step. This isolates a clear bottleneck and motivates more structured aggregation prompts (e.g.\ typed slots) as future work. Because every variable is recorded in the trace, the contradiction between \texttt{childers~=~"yes"} and \texttt{final~=~"Neither"} is directly verifiable.}
  \label{fig:fail-aggregation}
\end{figure}

%
%
%

\begin{figure}[t]
  \centering
  \begin{tcolorbox}[
      colback=pyrag-bg-bad, colframe=pyrag-bad,
      title={\textbf{Failure~F4}\hfill
      {\textcolor{pyrag-bad}{\ding{55}}~Incorrect}},
      fonttitle=\normalsize, boxrule=0.6pt, arc=2pt, left=4pt, right=4pt,
    ]
    \textbf{Question:} \emph{Are Northwestern University and Middlebury College both private schools?}\\[2pt]
    \textbf{Gold:} \texttt{yes} \hfill \textbf{Predicted:} \texttt{No}~\textcolor{pyrag-bad}{\ding{55}}
    \tcblower
\begin{lstlisting}[style=pyragcode]
schools = ["Northwestern University", "Middlebury College"]
flags   = {}
# Per-school predicate check -- both correctly resolved
for s in schools:
    ctx      = retrieve(f"Is {s} a private school?", topk=5)
    flags[s] = answer(f"Is {s} a private school?", ctx)
# flags = {"Northwestern University": "yes", "Middlebury College": "yes"}   [ok]
\end{lstlisting}
\begin{lstlisting}[style=pyragcode_warn]
# Final aggregation delegates the conjunction to the answer agent.
# The model's thinking flips both variables to "public" and returns "No".
final = answer(
    f"Given: Northwestern University is {flags[schools[0]]}, "
    f"Middlebury College is {flags[schools[1]]}. "
    f"Are Northwestern University and Middlebury College both private schools?",
    ctx,
)
# final -> "No"   [warn] contradicts flags == {...: "yes", ...: "yes"}
\end{lstlisting}
\begin{lstlisting}[style=pyragcode]
final = "yes" if all(v.lower() == "yes" for v in flags.values()) else "no"
\end{lstlisting}
  \end{tcolorbox}
  \caption{Boolean conjunction misexecuted by the answer agent. Both \texttt{flags} entries hold \texttt{"yes"}, but the answer agent's reasoning trace narrates ``\emph{both Northwestern University and Middlebury College are public institutions}'' before returning \texttt{"No"}. The conjunction is the failure point---precisely the operation Python expresses with one keyword. Replacing the final \texttt{answer(...)} call with the one-line \texttt{all(...)} expression shown above eliminates the failure mode entirely. This is the inverse of Case~C: when the boolean structure is enforced by the program rather than narrated to the LLM, the result is deterministic.}
  \label{fig:fail-bool}
\end{figure}

\begin{figure}[t]
  \centering
  \begin{tcolorbox}[
      colback=pyrag-bg-bad, colframe=pyrag-bad,
      title={\textbf{Failure~F5}\hfill
      {\textcolor{pyrag-bad}{\ding{55}}~Incorrect}},
      fonttitle=\normalsize, boxrule=0.6pt, arc=2pt, left=4pt, right=4pt,
    ]
    \textbf{Question:} \emph{Which of Aaron Goodwin's clients was born on May 31, 1984?}\\[2pt]
    \textbf{Gold:} \texttt{Nate Robinson} \hfill \textbf{Predicted:} \texttt{None}~\textcolor{pyrag-bad}{\ding{55}}
    \tcblower
\begin{lstlisting}[style=pyragcode]
ctx_cli = retrieve("Who are Aaron Goodwin's clients?", topk=5)
clients = answer("List Aaron Goodwin's clienhttps://gasolsun36.github.io/PyRAG/ts.", ctx_cli)
# clients is a *string*: "LeBron James, Dwight Howard, Chris Webber, ..."
\end{lstlisting}
\begin{lstlisting}[style=pyragcode_warn]
# The plan iterates `clients` directly. In Python, iterating a `str`
# yields one *character* per step -- not one client.
births = {}
for name in clients:                     # [warn] name is "L", then "e", "B", "r", ...
    ctx           = retrieve(f"When is the birthdate of {name}?", topk=5)
    births[name]  = answer(f"When is the birthdate of {name}?", ctx)
# Trace explodes: 400+ retrieve+answer calls, each on a single character.
# Most return "unknown"; a few accidentally hit "August 6, 1992", "October 13, 1992", ...
\end{lstlisting}
\begin{lstlisting}[style=pyragcode]
# Final aggregation cannot recover; no character matches "May 31, 1984"
final = answer(
    f"Given: {births}. Which client was born on May 31, 1984?",
    ctx,
)
# final -> "None"   [wrong]   (correct answer: Nate Robinson)
\end{lstlisting}
\begin{lstlisting}[style=pyragcode]
clients = [c.strip() for c in clients.split(",")]
\end{lstlisting}
  \end{tcolorbox}
  \caption{Type confusion in a \texttt{for}-loop. The bug is a single-line type confusion: \texttt{clients} is a comma-joined string, not a list, so iterating it character-by-character is silently legal Python and the executor fans out into hundreds of nonsensical retrievals on \texttt{"L"}, \texttt{"e"}, \texttt{"B"}, \dots~The fix is the one-line cast shown above, after which the original \texttt{for}-loop iterates over actual client names. Such failures are uniquely visible in the executable trace: the explosion of single-character queries makes the type error mechanically obvious, whereas in a free-form CoT the same confusion would surface as ``the model got distracted'' or ``hallucinated names.'' The failure mode is unique to the executable interface, but so is the diagnosis---the trace pinpoints the exact line that needs a \texttt{.split(",")}.}
  \label{fig:fail-typeconfusion}
\end{figure}